\documentclass[a4paper,fleqn]{cas-dc}
\usepackage[authoryear]{natbib} 

\usepackage{threeparttable}
\usepackage{flushend}
\usepackage{graphicx}

\usepackage{subcaption}

\def\tsc#1{\csdef{#1}{\textsc{\lowercase{#1}}\xspace}}
\tsc{WGM}
\tsc{QE}

\newdefinition{rmk}{Remark}
\usepackage{amsmath}
\usepackage{upgreek}
\usepackage{appendix}

\begin{document}
\begin{sloppypar} 

\let\WriteBookmarks\relax
\def\floatpagepagefraction{1}
\def\textpagefraction{.001}
\let\printorcid\relax

\makeatletter
\renewcommand{\fnum@figure}{Fig. \thefigure.\@gobble}
\makeatother

\shorttitle{}    

\shortauthors{}  

\title [mode = title]{ARMOR: Adaptive Meshing with Reinforcement Optimization for Real-time 3D Monitoring in Unexposed Scenes}

\tnotetext[1]{This research was supported by the National Natural Science Foundation Project ( No. 42201477, No.42130105)} 

\author[1]{Yizhe Zhang}

\ead{yizhezhang0418@whu.edu.cn}
\credit{Conceptualization of this study, Methodology, Software, Validation, Writing – original draft}
\affiliation[1]{organization={State Key Laboratory of Information Engineering in Surveying, Mapping and Remote Sensing},
            addressline={Wuhan University}, 
            city={Wuhan},
            postcode={430079}, 
            country={China}}

\author[2]{Jianping Li}
\cormark[1]
\ead{jianping.li@ntu.edu.sg}
\credit{Conceptualization of this study, Methodology, Writing review, Funding acquisition}
\affiliation[2]{organization={School of Electrical and Electronic Engineering},
            addressline={Nanyang Technological University}, 
            country={Singapore}}

\author[3]{Xin Zhao}

\ead{xinzhaodc@whu.edu.cn}
\credit{Conceptualization of this study, Maintenance, Validation}
\affiliation[3]{organization={School of Computer Science},
            addressline={Wuhan University}, 
            city={Wuhan},
            postcode={430079}, 
            country={China}}

\author[4]{Fuxun Liang}
\ead{liangfuxun@whu.edu.cn}
\credit{Conceptualization of this study, Data Collection, Methodology, Validation}
\affiliation[4]{organization={School of Urban Design},
            addressline={Wuhan University}, 
            city={Wuhan},
            postcode={430072}, 
            country={China}}

\author[1]{Zhen Dong}
\ead{dongzhenwhu@whu.edu.cn}
\credit{Conceptualization of this study, Methodology, Writing review, Project administration, Funding acquisition}

\author[1]{Bisheng Yang}
\ead{bshyang@whu.edu.cn}
\credit{Conceptualization of this study, Project administration, Funding acquisition}

\cortext[1]{Corresponding author}



\begin{abstract}
Unexposed environments, such as lava tubes, mines, and tunnels, are among the most complex yet strategically significant domains for scientific exploration and infrastructure development. Accurate and real-time 3D meshing of these environments is essential for applications including automated structural assessment, robotic-assisted inspection, and safety monitoring. Implicit neural Signed Distance Fields (SDFs) have shown promising capabilities in online meshing; however, existing methods often suffer from large projection errors and rely on fixed reconstruction parameters, limiting their adaptability to complex and unstructured underground environments such as tunnels, caves, and lava tubes.
To address these challenges, this paper proposes ARMOR, a scene-adaptive and reinforcement learning-based framework for real-time 3D meshing in unexposed environments. ARMOR consists of three primary components:
(1) a spatio-temporal geometry smoothing module that enhances point cloud density by leveraging temporal information and integrates a context-aware normal vector orientation strategy to improve SDF training in underground settings;
(2) a reinforcement learning-based parameter optimization module that dynamically adjusts reconstruction parameters in response to complex environmental characteristics; and
(3) ARMOR learns online parameter tuning strategies from prior point cloud datasets through simulation and practice, further improving its adaptability to unseen real environments.
The proposed method was validated across more than 3,000 meters of underground environments, including engineered tunnels, natural caves, and lava tubes. Experimental results demonstrate that ARMOR achieves superior performance in real-time mesh reconstruction, reducing geometric error by 3.96\% compared to state-of-the-art baselines, while maintaining real-time efficiency. The method exhibits improved robustness, accuracy, and adaptability, indicating its potential for advanced 3D monitoring and mapping in challenging unexposed scenarios. The project page can be found at: \href{https://yizhezhang0418.github.io/armor.github.io/}{https://yizhezhang0418.github.io/armor.github.io/}.

\end{abstract}





\maketitle

\begin{figure*}
    \centering
    \includegraphics[width=\linewidth]{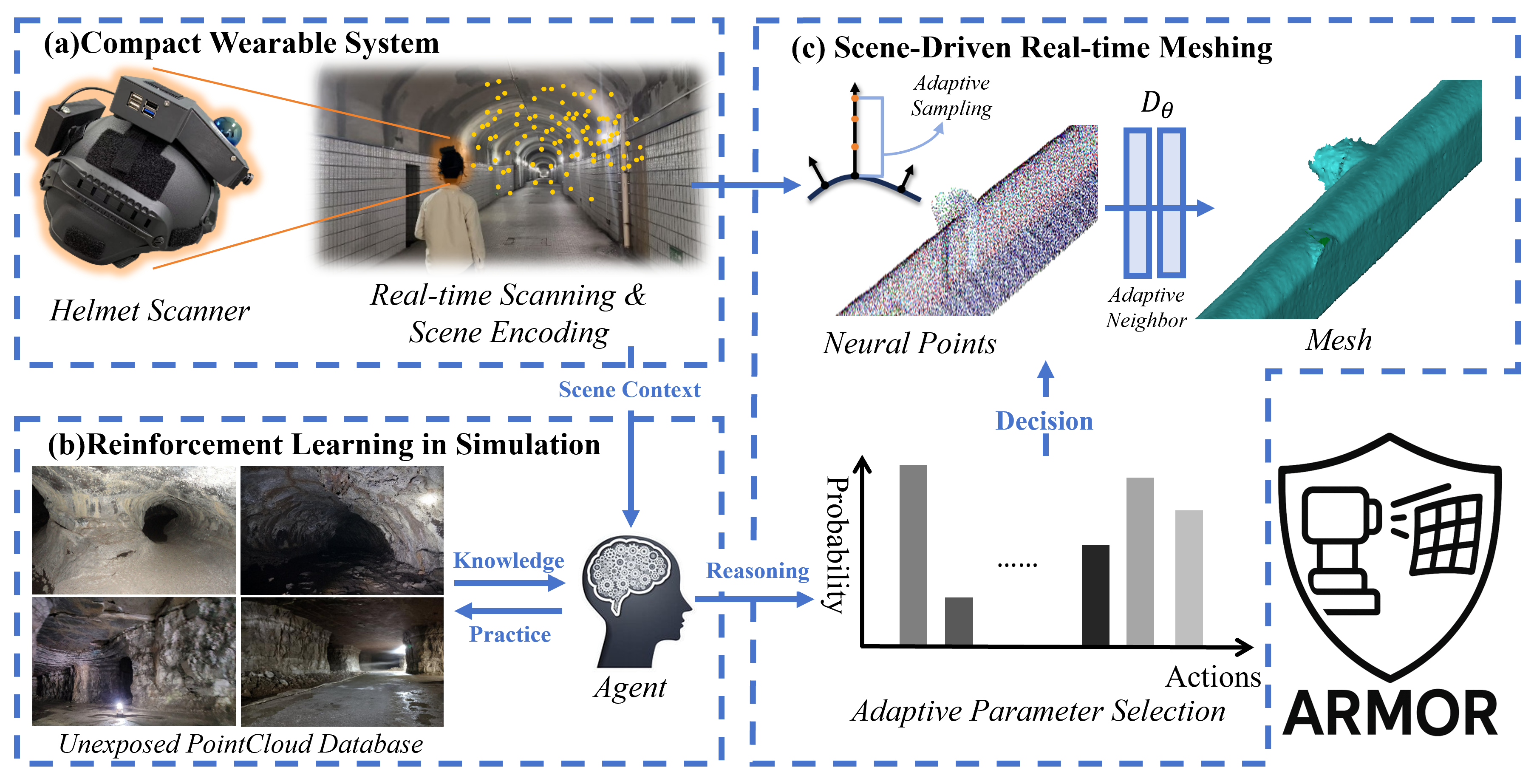}
    \caption{Adaptive Meshing with Reinforcement Optimization for Real-time 3D Monitoring in Underground Sites. Our system processes sequential LiDAR frames through three key stages: (1) spatial-temporal normal vector smoothing to enhance geometric consistency, (2) reinforcement learning-based parameter optimization that adapts to local scene characteristics, and (3) high-fidelity mesh reconstruction. This pipeline enables accurate and robust real-time 3D monitoring in complex underground environments, which is critical for construction safety and progress assessment.}
    \label{fig:main framework}
\end{figure*}

\section{Introduction}

Accidents in unexposed environments continue to pose a serious global threat, with underground incidents claiming thousands of lives each year~\citep{baraza2023statistical, dol_msha_2024}. In China alone, underground coal mine accidents account for around 100 fatalities in just six months~\citep{china_coal_2020}. A recurring factor in many of these tragedies is the absence of accurate, up-to-date 3D spatial information, which hampers timely rescue efforts and complicates risk assessment and decision-making.

Unexposed scenes, such as lava tubes \citep{martin2023pressurized}, mines \citep{si2021novel,lin2024graph}, and tunnels \citep{lin2006automated,zhang2024legged}, are among the most challenging yet strategically vital environments for modern scientific research and infrastructure construction, demanding precise 3D understanding under extreme observational constraints \citep{ebadi2023present}. These environments often exhibit complex geometries, limited accessibility, and poor lighting conditions, which pose significant challenges for accurate and efficient 3D reconstruction. Among existing technologies, LiDAR has emerged as a primary method for capturing the spatial structure of such environments \citep{li2023whu,lin2024graph, yang2024ubiquitous}. The resulting point clouds are typically processed offline to generate mesh models \citep{hoppe2009adaptive}, which serve as the foundation for downstream tasks such as structural integrity assessment, construction planning, and safety evaluation \citep{cui2023direct}.

Most current 3D meshing solutions follow a "scan-then-mesh" paradigm \citep{kolle2021hessigheim,cui2023direct}, where point clouds acquired in the field are later converted into mesh representations using post-processing techniques like Poisson surface reconstruction, ball-pivoting, and alpha shapes \citep{huang2024surface}. While these methods can produce high-quality meshes, they require extensive computational resources, expert parameter tuning, and significant processing time, often taking hours to complete for complex underground environments. This "scan-then-mesh" workflow and the time-consuming algorithms create a critical delay from data collection to actionable insights, severely limiting their applicability in time-critical tasks such as on-site safety monitoring and online structural inspection. Although existing methods have improved time efficiency, they often struggle to generate high-fidelity meshes at arbitrary resolutions, limiting their use in applications demanding detailed structural restoration \citep{lin2023immesh,ruan2023slamesh}.

Implicit Neural Fields (INFs) have emerged as a compelling alternative to traditional meshing methods, offering notable advantages in reconstructing continuous, high-fidelity scenes with significantly reduced processing time~\citep{schirmer2024geometric}. By encoding geometric information within neural networks, INFs enable precise surface reconstruction and have the potential to complete occluded or non-exposed regions. Through the generation of signed distance fields (SDFs), which capture the distance to the nearest surface, INFs can produce detailed mesh outputs using algorithms such as Marching Cubes~\citep{zhong2023shine}.
Despite their promise for real-time meshing, applying INFs to diverse unexposed environments remains highly challenging due to two primary factors. (1) the inherent noise and sparsity of real-time scanning data lead to unstable normal estimations, directly degrading the quality of the learned SDFs. Conventional SDF construction typically relies on projective distance metrics, which measure distances along LiDAR beams rather than true orthogonal distances to surfaces. This introduces systematic approximation errors that consistently overestimate surface proximity, significantly hindering accurate reconstruction in subterranean environments~\citep{wiesmann2023locndf}.  (2) INFs are highly sensitive to parameter settings, with optimal configurations varying not only across different sites but even within distinct regions of the same environment.  The high geometric variability characteristic of unexposed scenes means that parameters tuned for one area often perform suboptimally in others, creating a strong dependency on manual tuning and expert intervention.  These limitations frequently result in inconsistent reconstruction quality, manifested in geometric artifacts, incomplete surfaces, or over-smoothed structures, which can ultimately compromise the accuracy of spatial measurements and downstream analysis.

To tackle these challenges, we introduce ARMOR, a novel framework for accurate online meshing in unexposed environments, as illustrated in Fig.~\ref{fig:main framework}. ARMOR combines spatial-temporal geometry smoothing with reinforcement learning-based parameter adaptation to significantly improve online meshing quality and consistency. The main contributions of this paper are as follows:
\begin{enumerate}[1)]
\setlength{\leftmargin}{0em} 
\item A spatio-temporal fused normal smoothing paradigm that leverages multi-view consistency across sequential LiDAR frames to improve normal estimation stability and geometric coherence. By computing true distance fields instead of relying on projection-based metrics, this approach effectively eliminates SDF label errors and significantly enhances mesh quality, particularly in geometrically complex or sparsely observed regions.
\item A reinforcement learning framework for adaptive parameter optimization that autonomously selects optimal reconstruction parameters based on local scene characteristics. This eliminates the need for manual parameter tuning and ensures consistent reconstruction quality across diverse underground environments without human intervention.
\item  ARMOR learns online parameter tuning strategies from prior point cloud datasets through a combination of simulation and real-world deployment. This learning-based adaptation further enhances its robustness to unseen environments. Extensive evaluations on both synthetic underground benchmarks and real unexposed scenes demonstrate that ARMOR achieves superior reconstruction accuracy, completeness, and geometric fidelity.
\end{enumerate}

The rest of this paper is structured as follows. The related works are reviewed in Section \ref{Related works}. The preliminaries for our system are provided in Section \ref{Preliminary}. A detailed description of ARMOR is presented in Section \ref{Methodology}. The experiments are conducted on simulation and public and in-house helmet-based datasets in Section \ref{Experiment}. Conclusion and future work are drawn in Section \ref{Conslusion}.

\section{Related works}
\label{Related works}
\subsection{Traditional 3D Meshing Methods}

Traditional 3D meshing of unexposed scenes has primarily relied on offline post-processing of LiDAR point clouds. Algorithms such as the Ball-Pivoting Algorithm (BPA)~\citep{bernardini2002ball} and Poisson Surface Reconstruction~\citep{kazhdan2006poisson} have been widely adopted for underground modeling. While these techniques are capable of generating high-quality meshes, they typically require substantial computational resources and extended processing times—often taking several hours for complex underground structures~\citep{daroya2020rein}. In industrial applications, commercial software such as Leica Cyclone 3DR incorporates variants of these algorithms with domain-specific workflows tailored for tunnel analysis.


Recent advances in SLAM technology have facilitated real-time mapping solutions that partially overcome the latency constraints of traditional post-processed workflows. Modern LiDAR-based SLAM systems, for instance, are capable of dynamically reconstructing underground environments. However, these real-time approaches still encounter key limitations: they typically produce only sparse or semi-dense point clouds, which are inadequate for high-fidelity visualization or detailed geometric analysis~\citep{ebadi2020lamp,wang2022ulsm,trybala2023comparison}. Even when denser point clouds are acquired, generating high-quality mesh surfaces in real time remains computationally demanding, as surface reconstruction algorithms often require significant processing overhead. As a result, most current SLAM frameworks produce incomplete or low-resolution mesh models that lack the geometric detail necessary for precise measurement and structural analysis in unexposed scenarios.
Although some SLAM systems are capable of generating mesh outputs directly, these meshes frequently suffer from insufficient fidelity to accurately represent the complex geometries typical of unexposed environments~\citep{ruan2023slamesh,lin2023immesh}. This fundamental limitation underscores the pressing need for advanced real-time meshing techniques that can intelligently refine and densify geometry while preserving structural accuracy.

\subsection{Implicit Neural Fields for 3D Meshing}
Implicit neural fields have emerged as a powerful paradigm for 3D representation, offering continuous and high-fidelity scene modeling. Neural Radiance Fields (NeRF)~\citep{mildenhall2021nerf} pioneered this direction by using neural networks to encode view-dependent appearance and 3D geometry for novel view synthesis. This approach demonstrated exceptional capabilities in representing complex scenes with high visual fidelity. Subsequent works explored implicit geometric representations through Signed Distance Functions (SDFs)~\citep{park2019deepsdf, chibane2020implicit} and occupancy fields~\citep{mescheder2019occupancy}, which enabled direct mesh extraction through methods like Marching Cubes~\citep{lorensen1998marching}.

The integration of implicit neural fields with SLAM systems marked a significant advancement in real-time 3D reconstruction. iMAP~\citep{sucar2021imap} first demonstrated the feasibility of using implicit neural representations for incremental mapping, while NICE-SLAM~\citep{zhu2022nice} extended this approach with a hierarchical feature grid structure to improve reconstruction quality and efficiency. In the meanwhile, Vox-Fusion~\citep{yang2022vox} adapt an octree structure to implement a dynamic voxel allocation strategy. GO-SLAM~\citep{zhang2023go} further enhanced these capabilities by utilization of learned global geometry from input history. These systems, however, faced challenges in computational efficiency and reconstruction detail when applied to complex environments like underground tunnels.


More recently, point-based implicit neural mapping approaches have gained prominence for their ability to model fine-grained geometric details while maintaining computational efficiency. Point-SLAM~\citep{sandstrom2023point} leverages a sparse set of adaptive 3D points with associated neural features to represent scenes, significantly improving reconstruction quality. NeRF-LOAM~\citep{deng2023nerf} pioneered the use of purely LiDAR-based incremental odometry and mapping, employing an octree structure to maintain voxel maps and hash grids to accelerate search operations, thereby enhancing computational efficiency. LONER~\citep{isaacson2023loner} adopts an architecture similar to Instant-NGP~\citep{muller2022instant}, encoding scenes in a coarse-to-fine manner through multi-resolution hash encoding to achieve real-time performance. PIN-SLAM~\citep{pan2024pin} further advances this methodology by utilizing point-based representations with interpolated neural features, enabling both accurate camera tracking and high-fidelity surface reconstruction. These point-based approaches show particular promise for underground environments due to their ability to represent complex surfaces while maintaining real-time performance. Our previous work~\citep{zhang2024nerf} has addressed aspects of dense underground model reconstruction; however, significant challenges remain regarding parameter sensitivity and normal estimation accuracy, especially in complex underground structures with irregular geometry and variable surface properties.

\subsection{Reinforcement Learning for Adaptive Parameter Optimization}
Reinforcement Learning (RL) has demonstrated significant potential for solving complex parameter optimization problems across various domains. For example, RL has been successfully employed in computer vision for hyperparameter tuning~\citep{liu2024scale,wang2024self}, in SLAM for a more robust tracking ~\citep{messikommer2024reinforcement}, in robot navigation for being cross in a crowded environment ~\citep{cao2024learning}. Despite these advances, the application of RL in the context of implicit neural field reconstruction remains largely unexplored. Existing 3D reconstruction methods for unexposed environments, such as PIN-SLAM ~\citep{pan2024pin} and N\textsuperscript{3}-Mapping ~\citep{song2024n}, usually rely on fixed or manually tuned parameters. PIN-SLAM suffers from projection errors due to using training labels derived from projective distance measurements, while N\textsuperscript{3}-Mapping, although incorporating normal-guided sampling, fails to achieve real-time performance and it is hard to get robust normal estimation in a complex unexposed scene. Consequently, these fixed-parameter and with inherent error approaches often meet inconsistent reconstruction quality, such as artifacts and information loss. In this work, we bridge this gap by introducing a reinforcement learning-based parameter adaptation strategy tailored for implicit neural reconstruction. Our method leverages RL agent to dynamically adjust meshing parameters according to local scene characteristics, thereby enhancing reconstruction consistency and accuracy in challenging underground environments.

\section{Preliminary for Reinforcement Learning}
\label{Preliminary}



Modern SLAM systems with implicit neural representations involve numerous parameters that significantly impact reconstruction quality in complex unexposed environments.  Manual parameter tuning becomes impractical across diverse scenarios. Reinforcement Learning (RL) addresses this by modeling parameter selection as a sequential decision process, where an agent learns optimal strategies through environmental feedback~\citep{hu2020voronoi, li2017deep}. This process is formalized as a Markov Decision Process (MDP), formalized as a tuple $(\mathbf{S}, \mathbf{A}, \mathbf{P}, \mathbf{R}, \gamma)$, where $\mathbf{S}$ represents the state space, $\mathbf{A}$ denotes the action space, $\mathbf{P}$ defines transition probabilities, $\mathbf{R}$ specifies the reward function, and $\gamma \in [0,1)$ represents the discount factor. The agent aims to find an optimal policy ${\pi^*}$ that maximizes the expected cumulative discounted reward:
\begin{equation}
\mathcal{J}(\mathbf{{\pi}}) = \mathbb{E}_{\boldsymbol{\tau} \sim {\pi}} \left[ \sum_{t=0}^{\infty} \gamma^{t} r_{t} \right],
\label{discounted_rewards}
\end{equation}
where $\boldsymbol{\tau}$ represents the trajectory of states, actions, and rewards, $t$ represents timestamp, and $\mathbb{E}_{\boldsymbol{\tau} \sim {\pi}}(\cdot)$ denotes expectation over trajectories induced by policy ${\pi}$.

\subsection{Actor-Critic}

Actor-Critic algorithms combine policy-based and value-based reinforcement learning to optimize agent performance. The framework consists of two components: the actor, which selects actions according to a parameterized policy ${\pi_{\boldsymbol{\theta}}}$, and the critic, which evaluates the value function $V(\mathbf{s}_t)$. The actor aims to maximize the expected cumulative reward, updating the policy using the policy gradient:
\begin{equation}
\nabla_{\boldsymbol{\theta}} J(\pi_{\boldsymbol{\theta}}) = \mathbb{E}_t \left[ \nabla_{\boldsymbol{\theta}} \log \pi_{\boldsymbol{\theta}}(\mathbf{a}_t | \mathbf{s}_t) \hat{A}_t(\mathbf{s}_t, \mathbf{a}_t) \right],
\end{equation}
where $\hat{A}_t(\mathbf{s}_t, \mathbf{a}_t)$ is the advantage function, quantifying how much better an action is compared to the average in state $\mathbf{s}_t$. The critic computes the temporal difference (TD) error:
\begin{equation}
    \delta_t = r_t + \gamma V(\mathbf{s}_{t+1}) - V(\mathbf{s}_t).
\end{equation}
By providing a more stable estimate of the advantage function, the critic reduces variance in the policy gradient updates and stabilizes the learning process.

\subsection{Proximal Policy Optimization for Actor-Critic}
While Actor-Critic is effective, it may suffer from instability in the updates, especially when the policy update is great. Proximal Policy Optimization (PPO) is put up to address the problem by stabilizing the policy updates in the Actor-Critic method~\citep{schulman2017proximal}. PPO is a policy optimization method that modifies the standard policy gradient approach by introducing a clipped surrogate objective. The idea is to ensure the new policy doesn't deviate too far from the old policy, even if the objective function suggests a large update.

In PPO, the objective function for the actor is changed to use the clipped surrogate approach. The clipped objective function in PPO is defined as:
\begin{equation}
L^{\rm{CLIP}}(\boldsymbol{\theta}) = \hat{\mathbb{E}}_t \left[ \min \left( r_t(\boldsymbol{\theta}) \hat{A}_t, \textit{clip} \left( r_t(\boldsymbol{\theta}), 1-\epsilon, 1+\epsilon \right) \hat{A}_t \right) \right],
\end{equation}
where $r_t(\boldsymbol{\theta})$ is the probability ratio between the new policy and the old policy, $\epsilon$ is the hyperparameter that controls the clipping range, and $ \textit{clip} \left( r_t(\boldsymbol{\theta}), 1-\epsilon, 1+\epsilon \right)$ is the clipped ratio that limits how much the probability ratio can change. The clip method limits the change in the policy and prevents overly large updates.

In summary, the theoretical progression from fundamental reinforcement learning principles to Actor-Critic architectures and ultimately to PPO reflects a systematic advancement in policy optimization methodology. While Actor-Critic algorithms combine policy and value estimation to reduce variance, they can still suffer from instability due to excessively large policy updates. PPO addresses this limitation by introducing a clipped surrogate objective, which directly constrains the magnitude of policy updates at each iteration. This clipping mechanism ensures that the new policy does not deviate excessively from the previous policy, thereby promoting stable and monotonic policy improvement. As a result, PPO achieves more reliable convergence, particularly in the complex and geometrically diverse environments characteristic of unexposed reconstruction tasks.

\begin{figure*}[htb]
\centering
\includegraphics[width=.95\linewidth]{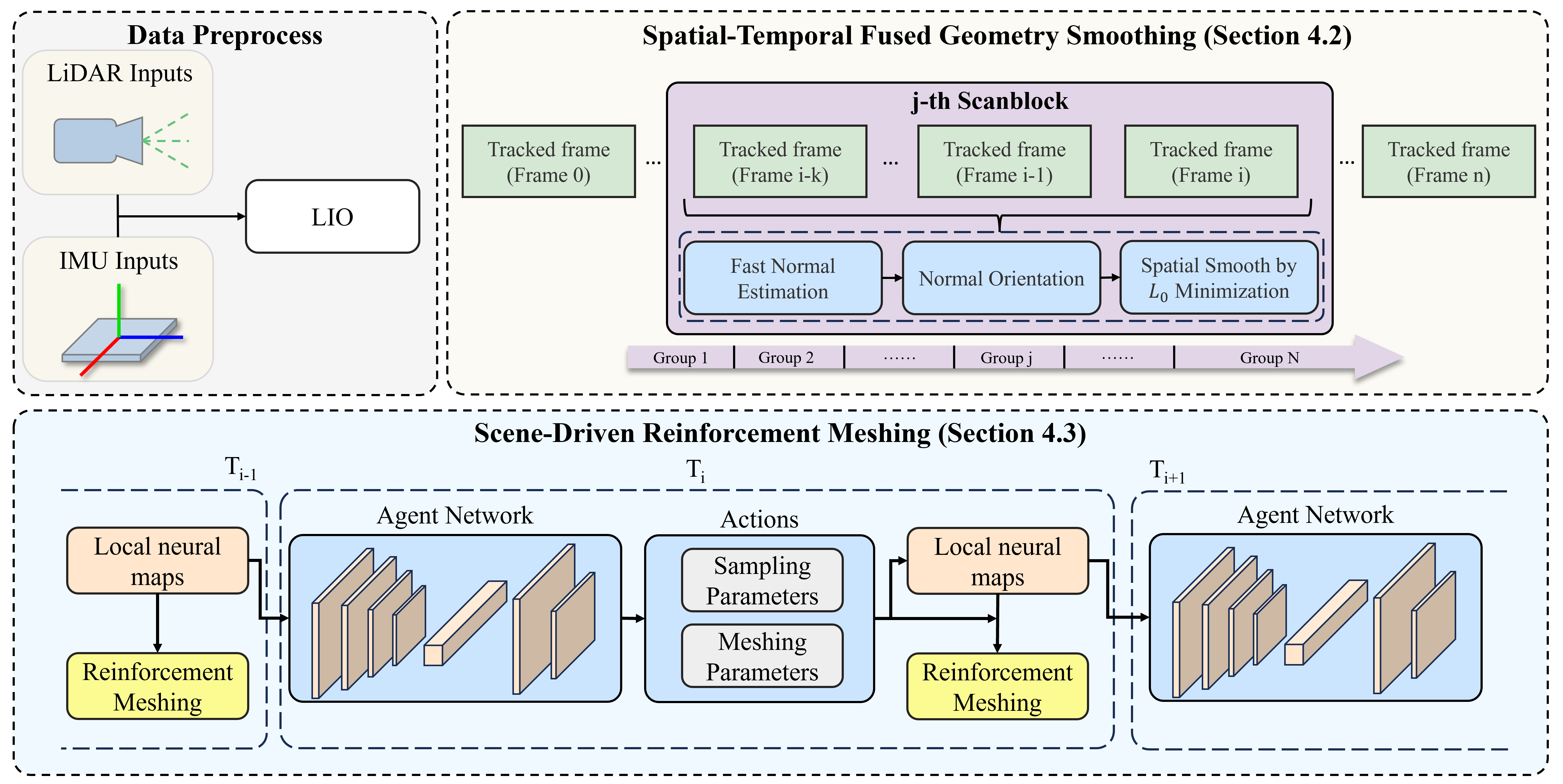}
\caption{Architecture of ARMOR. The ARMOR pipeline begins with the preprocessing of sequential LiDAR and IMU data, followed by a spatio-temporal geometry smoothing module that leverages consistency constraints to generate high-quality normal estimations. A reinforcement learning agent then analyzes the characteristics of local neural maps from the previous state $T_{i-1}$ and computes a multi-discrete probability distribution to infer optimal mapping parameters. These parameters guide the update of the local neural map and drive the adaptive meshing process. The updated neural representation is then used as the observation for the next timestamp $T_{i+1}$, forming a continuous loop for adaptive reconstruction.}
\label{pipeline}
\end{figure*}

\section{Our Approach: ARMOR}
\label{Methodology}
In this paper, we propose \textbf{ARMOR}, an adaptive online meshing framework tailored for unexposed environments. The overall workflow is illustrated in Fig.~\ref{pipeline}. Given asynchronous LiDAR and high-frequency IMU inputs, we first accumulate point cloud data and perform robust LiDAR-Inertial Odometry (LIO). When a loop closure is detected, we employ Iterative Closest Point (ICP)~\citep{besl1992method} to compute the closure constraint, which is then fused with front-end odometry using GTSAM~\citep{gtsam} for factor graph optimization, yielding more consistent pose estimates.
The estimated poses and accumulated point clouds are subsequently synchronized and processed via a spatio-temporal geometry smoothing module (Section~\ref{STFGS}), which enhances surface normal stability and geometric consistency. In the back-end meshing thread, the scene is represented as a collection of learnable neural points, where accurate Signed Distance Function (SDF) labels are derived from the smoothed geometry. A reinforcement learning agent is then employed to extract local features and predict optimal reconstruction parameters using a neural network (Section~\ref{SDRM}).
Throughout the SLAM process, the system is capable of real-time mesh reconstruction using the Marching Cubes algorithm, enabling continuous and adaptive 3D mapping of unexposed environments.

\subsection{Spatial-Temporal Fused Geometry Smoothing} \label{STFGS}
\begin{figure}[htb]
\centering
\includegraphics[width=\linewidth]{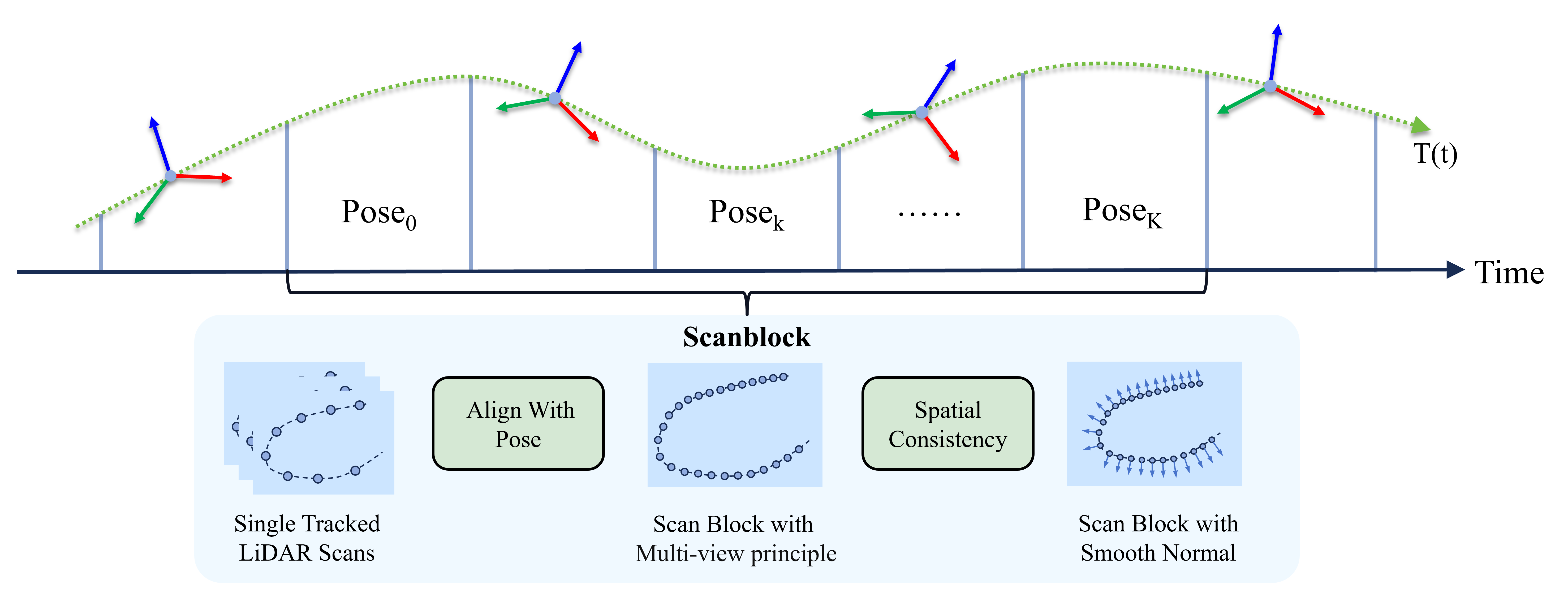}
\caption{Scanblock formation process. Trajctory \textbf{T}(t) shows the continuous movement of LiDAR over interval [$t_0,t_K$). In each interval, consecutive scan frames will integrate through coordinate transformation, enhancing point cloud density and geometric completeness for improved normal estimation in the unexposed environment.}
\label{scanblock}
\end{figure}

Using multi-view observation, we propose a novel data structure, termed the \textbf{\textit{scanblock}}, which segments point cloud frames tracked by the front-end into temporal blocks based on fixed time intervals. By aggregating point clouds over a specified duration, scanblocks effectively densify spatial observations and enhance geometric fidelity. As illustrated in Fig.~\ref{scanblock}, the scanblock thread continuously accumulates a sequence of scan frames, where each scan frame corresponds to point clouds collected within a 0.1-second window.
 
We denote the pose of the \(i\)-th frame as \(\mathbf{T}_i\), and its corresponding point cloud measurements are represented as \(\mathbf{C}_i\). A scan block, denoted as \(\mathbf{B}_j\), is composed of \(k\) adjacent scan frames. For each scan block \(\mathbf{B}_j\), we designate the pose of the first frame as the pose of the scan block. The point clouds of the subsequent \(k-1\) frames are then transformed using their respective poses, forming a complete scan block, as expressed in Eq. (\ref{scanblock_formula}).
\begin{equation}
    \label{scanblock_formula}
    \mathbf{B}_j = \mathbf{C}_0 \oplus \sum_{i=1}^{k-1} (\mathbf{T}_0^{-1} * \mathbf{T}_i * \mathbf{C}_i),
\end{equation}
where $\oplus$ indicates the concatenate operation.
Subsequently, we perform geometry smoothing on the point cloud normals within the scan block to achieve more consistent spatial geometric structures. 

To effectively capture the complex geometric structures of unexposed environments, we compute initial normal vectors \(\hat{\mathbf{n}}\) for each point using Principal Component Analysis (PCA)~\citep{mackiewicz1993principal} based on its $k$-nearest neighbors within a defined radius. To address the unique challenges of normal vector orientation in unexposed spaces, where traditional visibility-based methods often fail, we introduce a \textbf{M}ulti-\textbf{S}egment \textbf{C}entroid-based \textbf{N}ormal \textbf{V}ector \textbf{O}rientation Method (MSC-NVO). This approach divides the scanblock into multiple segments and uses their spatial relationships to establish consistent normal vector directions, significantly improving orientation accuracy in complex unexposed environment structures.


Specifically, we calculate the bounding box of a single scan block \(\mathbf{B}_j\). The orientation of the longest edge of the bounding box is treated as the primary direction, and the entire scan block is partitioned into \(N_{sg}\) segments along this direction. Next, the centroid of each segment is computed, and the centroids are connected to form a centroid line. For each point \(\mathbf{p}_i\) within the scan block, we calculate its projection \(\mathbf{proj}_i\) onto the corresponding segment of the centroid line. The direction of the normal vector \(\mathbf{n}_i\) is then determined by:

\begin{equation}
\mathbf{n}_i=
\begin{cases}
    \mathbf{n}_i & \text{if } \mathbf{n}_i \cdot \dfrac{\mathbf{proj}_i}{\Vert\mathbf{proj}_i\Vert} \geq 0, \\
    -\mathbf{n}_i & \text{if } \mathbf{n}_i \cdot \dfrac{\mathbf{proj}_i}{\Vert\mathbf{proj}_i\Vert} < 0.
\end{cases}
\label{normalorientation}
\end{equation}

To achieve a more reliable geometric description, inspired by PSS-BA~\citep{li2024pss}, we incorporate \(L_0\) minimization to smooth the point cloud normals using a spatial smoothing kernel. The optimization is performed based on Eq. (\ref{min_l0norm}):

\begin{equation}
    \label{min_l0norm}
    F = \min _{\mathbf{n},\left|\mathbf{n}\right|=1} (1-\mathbf{n}^{\top}\mathbf{\hat{n}}) + \eta\lvert \mathbf{D}(\mathbf{n})\rvert_{0},
\end{equation}
where \( \mathbf{D}(\mathbf{n})_{ik+j} = \mathbf{n}_i - \mathbf{n}_{\mathbf{N}(i,j)} \) and \( \mathbf{N}(i,j) \) denotes index of the \(j\)-th nearest neighbor of point \(i\) in the point cloud, \(k\) represents the total number of nearest neighbors considered for each point, and \(\lvert\circ\rvert_0\) represents the number of non-zero elements in the matrix. The first term of this equation aims to preserve data fidelity as much as possible, while $\eta$ controls the degree of smoothing. By minimizing the $L_0$ norm of the normal differences, our approach effectively suppresses noise and outliers in the estimated normals, leading to more consistent and reliable normal vectors across the point cloud. This leads to higher-quality geometric representations, which are crucial for downstream tasks such as surface reconstruction and mesh generation in complex unexposed environments.

\subsection{Scene-Driven Reinforcement Meshing}\label{SDRM}
In this section, we provide a comprehensive explanation of how the proposed approach enables the generation of dense and accurate meshes through interaction with the environment. In Section~\ref{Scene Representation}, we introduce our point-based implicit neural field representation and detail how neural points are trained using normal-guided sampling to eliminate projection errors, thereby facilitating more reliable meshing of unexposed environments. In Section~\ref{State Representation and Action Space}, we formulate the mapping task as a Markov Decision Process (MDP) and define the state representation derived from local neural point clouds, along with the multi-dimensional action space encompassing critical reconstruction parameters. Section~\ref{RL_network} presents our specialized network architecture that leverages sparse convolutions to efficiently process irregular neural point distributions while capturing multi-scale features. Finally, in Section~\ref{Reward Design}, we explain our composite reward function that combines multiple geometric quality metrics to guide the reinforcement learning agent toward optimal parameter selection across diverse unexposed environments. Together, these components form an adaptive reconstruction framework that dynamically adjusts meshing parameters based on local scene characteristics.


\subsubsection{Scene Representation}
\label{Scene Representation}

\begin{figure}[!htbp]
    \centering
    \includegraphics[width=0.95\linewidth]{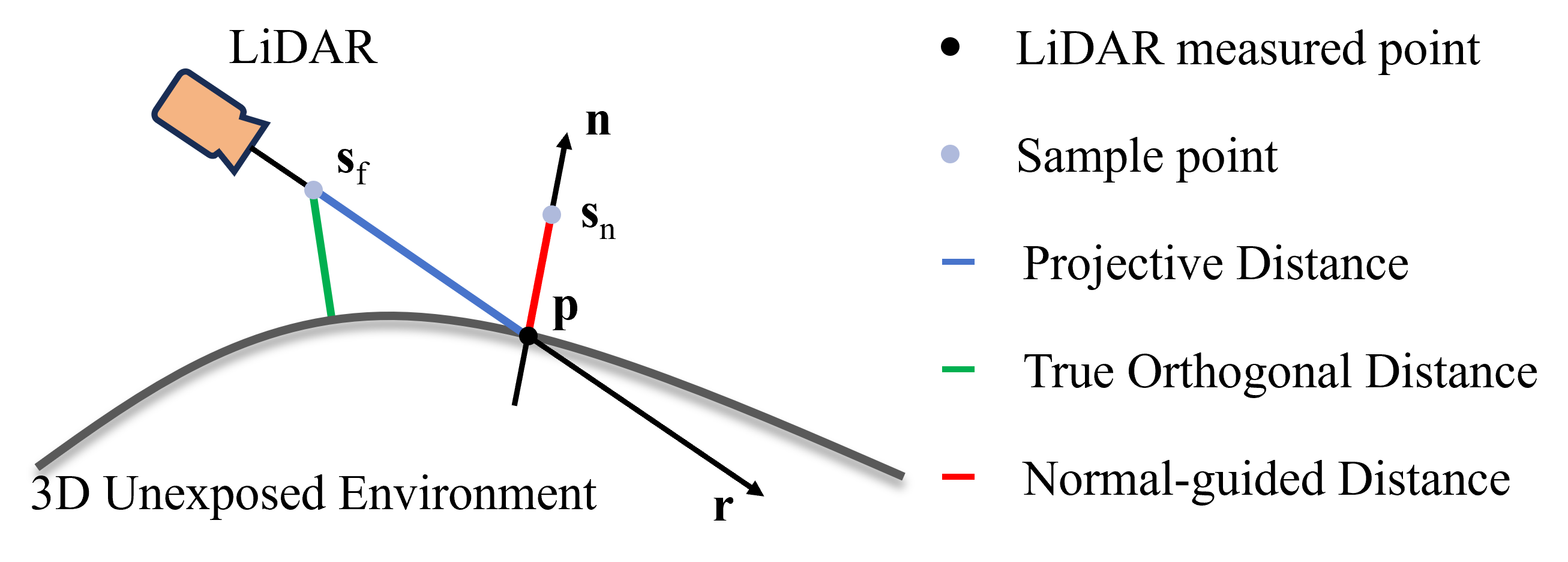}
    \caption{Illustration of our normal-guided sampling strategy for improved reconstruction. Projective distances along the LiDAR ray can introduce inherent SDF label errors in irregular 3D unexposed environments. By integrating enhanced geometry smoothing with normal-guided sampling, our method enables more reliable SDF labels and improves reconstruction quality.}
    \label{fig:sample_distance}
\end{figure}

We employ a point-based neural implicit representation with an auto-decoder architecture~\citep{park2019deepsdf}, predicting the SDF value at query position $\mathbf{x}$ conditioned on $K$-neighborhood neural points. For field training, we utilize normal-guided sampling to address projection bias, extracting $N_s$ sample points along surface normal vectors $\mathbf{n}$ and computing true signed distances as unbiased SDF labels, as shown in Fig.~\ref{fig:sample_distance}. 

To minimize ambiguity, we enforce a truncation region $[-tr, tr]$ near surfaces where samples follow a Gaussian distribution $\mathcal{N}(0, \sigma_s^2)$, with $\sigma_s$ controlling sampling density. To reduce artifacts in incremental mapping process, we further sample $N_f$ points uniformly in free space along the ray between sensor origin and the truncated space with uniform distribution \(U(\eta_{min}\Vert\mathbf p\Vert_2, \eta_{max}\Vert\mathbf p\Vert_2)\), while \(\Vert\mathbf p\Vert_2\) is the distance between sensor origin and LiDAR measured object, \(\eta_{min}\) and \(\eta_{max}\) stand for the ratio of the sampling region along the LiDAR ray. The field is optimized by minimizing a weighted combination of BCE and Eikonal losses, enabling online neural map construction. Mesh extraction via marching cubes is performed when the neighboring neural point count exceeds the threshold $N_{nn}$, ensuring adequate geometric support.

\subsubsection{State Representation and Action Space}
\label{State Representation and Action Space}

We formulate our real-time adaptive meshing task as MDP, and define the state representation as a feature embedding derived from neural point observations. The mapping thread processes tracked scanblocks, each comprising an estimated pose and associated point cloud data. Upon receiving a scanblock, the system encodes the point cloud into a set of neural points with latent features. To maintain computational efficiency without sacrificing reconstruction quality, we apply uniform voxel downsample to the local neural point map. The subsampled points are then processed through a feature extraction network to generate a compact feature embedding that constitutes the system state $\mathbf{s}_t$.

Following state-action pair $(\mathbf{s}_t, \mathbf{a}_t)$, the system transitions to subsequent state $\mathbf{s}_{t+1}$ according to transition probability $p_t \in \mathbf{P}$, receiving immediate reward $r_t \in \mathbf{R}$ attenuated by discount factor $\gamma$ for future timesteps. During operation, when the agent receives a state embedding, the policy $\mathbf{\pi}$ generates a probability distribution over possible actions. The final action selection is determined through a maximum likelihood strategy, selecting the action with the highest probability according to the current policy $\mathbf{\pi}$.

The action space $\mathbf{A}$ of our parameter optimization framework is six-dimensional, encompassing critical parameters that govern the neural implicit field training and mesh extraction processes:
\begin{equation}
    \label{action}
    \mathbf{a}_t = \left[\sigma_s, N_s, N_f, \eta_{min}, \eta_{max}, N_{nn}\right] \in \mathbf{A},
\end{equation}
where $\sigma_s$ controls the sample point distance distribution, $N_s$ and $N_f$ determine the number of surface and free-space sample points respectively, $\eta_{min}$ and $\eta_{max}$ define the sampling ratio bounds along sensor rays, and $N_{nn}$ specifies the minimum neural point support threshold required for mesh reconstruction. This parameterization enables fine-grained control over the reconstruction quality-efficiency trade-off across diverse environmental conditions.

\subsubsection{Reinforcement Meshing Network}
\label{RL_network}

\begin{figure}
    \centering
    \includegraphics[width=0.95\linewidth]{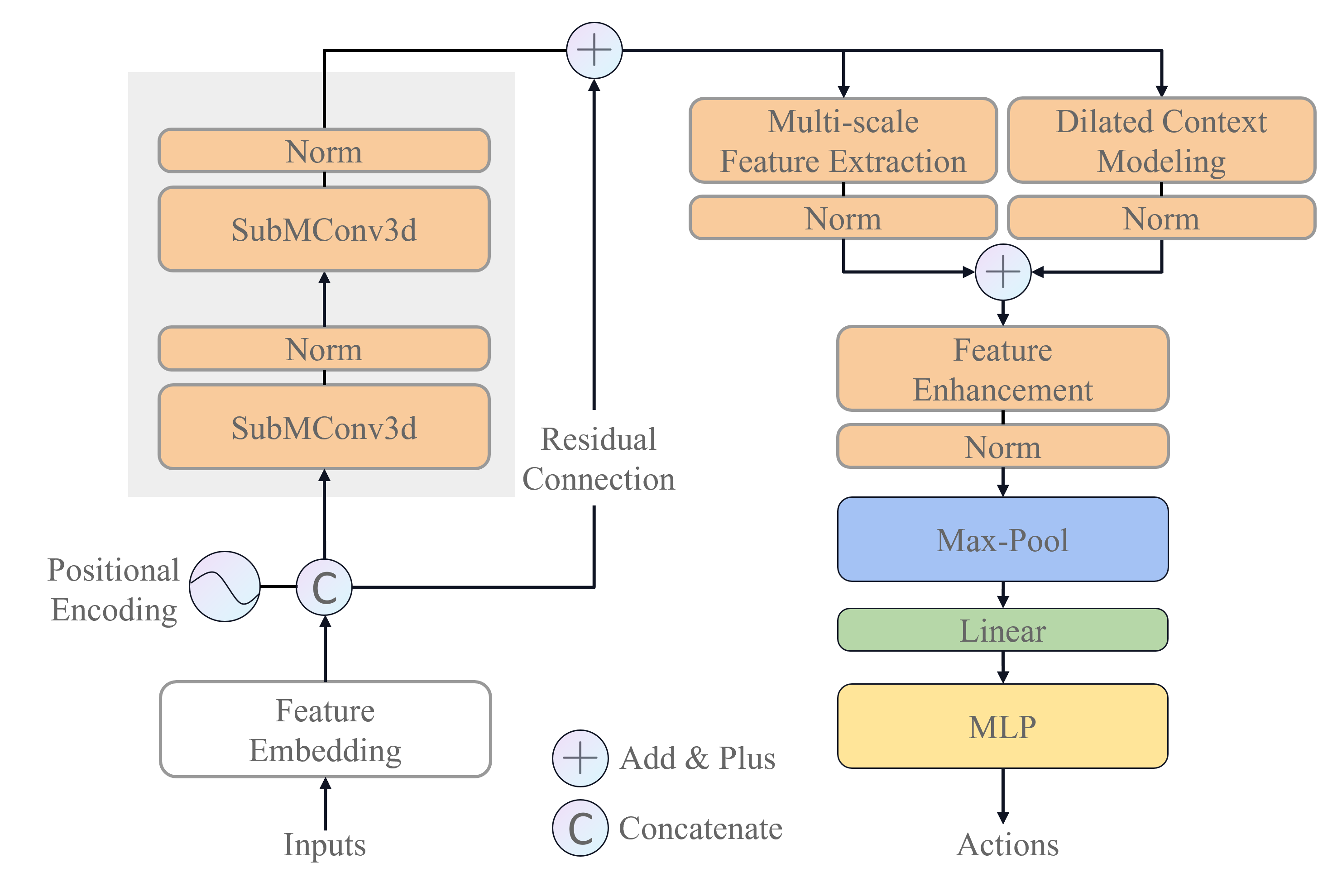}
    \caption{The architecture of agent network.}
    \label{fig:network}
\end{figure}

In this chapter, we give a detail description of our reinforcement meshing network. The network architecture we designed integrates hierarchical sparse convolutions, residual learning modules, and multi-scale feature extraction strategies to effectively process point cloud data from complex unexposed environments. As illustrated in Fig.~\ref{fig:network}, our architecture centers on a feature encoder that processes local map information. This encoder comprises multiple specialized sparse convolutional layers arranged in complementary configurations: one series of layers employs progressive downsampling convolutions, systematically expanding the receptive field to capture broader contextual information, while the other set utilizes dilated convolutions to maintain spatial resolution while enhancing feature discrimination capability without increasing parameter count. This architectural design enables simultaneous extraction of fine-grained geometric details and global structural patterns critical for accurate unexposed environment reconstruction. The network outputs action probabilities as multiple independent categorical distributions, allowing the agent to select optimal parameters for each dimension of the reconstruction process. During operation, the downsampled neural point cloud is processed through the encoder to extract latent features, which are then fed into a two-layer MLP. Final parameter selection occurs through maximum likelihood sampling. This approach enables dynamic adaptation to varying scene complexities while maintaining computational efficiency, critical for real-time unexposed environment reconstruction.

\rmk{
It is important to highlight that, given the complexity of unexposed environments and the constraints of real-time computation, our reinforcement learning framework requires a network architecture capable of efficiently handling variable-sized neural point inputs across diverse underground scenes. To address this, we adopt sparse convolutional networks as the backbone for feature extraction due to three key advantages: their computational efficiency when processing highly sparse point distributions typical of underground environments; their reduced memory overhead achieved by performing computations only at non-zero feature locations; and their ability to preserve irregular geometric structures that are often compromised during conventional voxelization.
}

\subsubsection{Reward Design}
\label{Reward Design}

The reward function is designed to guide parameter selection during the reconstruction process, enabling the agent to converge toward optimal geometric fidelity in challenging unexposed environments. We formulate the reward signals to directly correspond to well-established quantitative metrics that objectively assess reconstruction quality. This ensures that improvements in the policy directly translate into measurable gains in reconstruction performance.
To this end, we define a composite reward signal as a weighted linear combination of four complementary quality metrics:

\begin{equation}
r_{sum} = \lambda_1 \cdot r_{acc} + \lambda_2 \cdot r_{comp} + \lambda_3 \cdot r_{cham} + \lambda_4 \cdot r_{F},
\end{equation}
where the $\lambda_1, \lambda_2, \lambda_3, \lambda_4$ are hyperparameter that control the relative importance of each quality metrics in the overall reward signal.

For the distance-based metrics (accuracy, completeness, and Chamfer distance), we implement a consistent piecewise linear mapping function that transforms raw geometric error measurements (in centimeters) into normalized reward signals. This transformation provides positive reinforcement for high-quality reconstructions while imposing graduated penalties as reconstruction errors increase, creating a well-shaped reward landscape that facilitates efficient learning.

\textbf{Accuracy Reward ($r_{acc}$)}: Quantifies the geometric precision of the reconstruction by measuring the average distance from points on the reconstructed mesh to their nearest counterparts in the ground truth point cloud.

\textbf{Completeness Reward ($r_{comp}$)}: Evaluates reconstruction coverage by measuring the average distance from ground truth points to their nearest neighbors on the reconstructed mesh, similarly mapped to reward values.

\textbf{Chamfer Reward ($r_{cham}$)}: Combines accuracy and completeness metrics to provide a comprehensive measure of geometric similarity between reconstructed and ground truth surfaces.

For the F-score metric, which differs in nature from distance-based errors, we apply a separate mapping function, which converts the F-score signal to reward values, and higher F-scores is represented as greater positive reward.

\textbf{F-score Reward ($r_{F}$)}: Incorporates a classification-based evaluation approach by calculating the harmonic mean of precision and recall at a fixed distance threshold (15cm), transformed through a dedicated mapping function that rewards higher F-scores with greater positive values.

This multi-objective reward formulation effectively balances various aspects of reconstruction quality, enabling the agent to learn parameter adaptation strategies that optimize the geometric fidelity of reconstructed meshes across diverse unexposed environments while maintaining computational efficiency.

\section{Experiment}
\label{Experiment}
The primary focus of this work is to integrate geometry in the spatial-temporal space and utilize an agent to select optimal parameters for adaptive reconstruction based on the scene characteristics. In this section, we analyze the capabilities of our approach and evaluate its performance. All experimental methods are implemented in C++ and Python, and all experiments are conducted on an Intel® Core i7-14700KF CPU and an Nvidia® GeForce RTX 4090 GPU.

\subsection{Experimental Setup}

In our spatial-temporal smoothing framework, we empirically set the scanblock size to 20 consecutive frames to balance computational efficiency and temporal consistency. For geometry smoothing, we employ a spherical neighborhood search with a radius of 2 meters and limit the maximum number of neighboring points to 20 for robust normal estimation. The smoothing weight $\beta$ is set to 1.0 to ensure fast convergence of the optimization process, while the preservation weight $\eta$ is set to 0.1 to maintain fine geometric details while removing noise. These parameters were determined through extensive experimentation to achieve an optimal trade-off between smoothness and detail preservation in unexposed environments. 

\subsection{Learning and Practice in Simulation}
\begin{figure}[!htbp]
    \centering
    \includegraphics[width=0.95\linewidth]{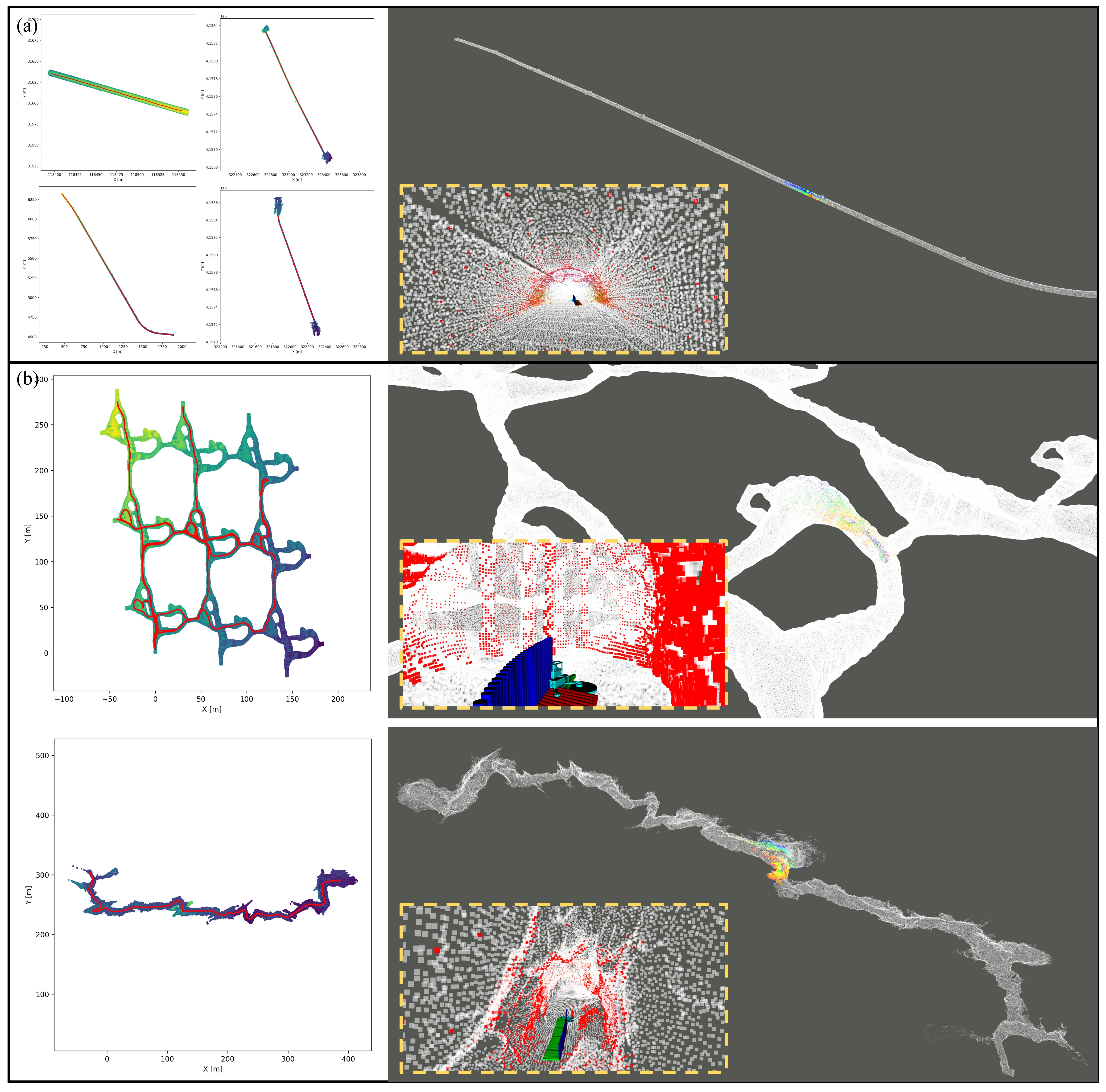}
    \caption{Representative subset of the simulation database used for learning and practice. The illustrated examples cover two primary categories: (a) regular tunnel environments and (b) complex cave environments. Red trajectories in the left panels depict the UAV flight paths, while the right panels illustrate the corresponding LiDAR scanning patterns within each simulated environment.}
    \label{fig:database}
\end{figure}

To enable our agent to generalize across various scenarios, we designed different unexposed training environments in MARSIM~\citep{kong2023marsim}, including complex unexposed tunnels and unexposed caves, as partially illustrated in Fig.~\ref{fig:database}. The LiDAR used for simulation is Livox Mid-360~\citep{Livox}, a low-cost LiDAR that is widely used for 3D mapping. In unexposed tunnels, we use the data collected from real world. For example, crossing tunnel in urban~\citep{jeong2019complex, autowarefoundation}, which primarily focuses on tunnel construction applications and by the need for high reconstruction accuracy to faithfully capture intricate structural details. In unexposed caves, we use the data both from real world~\citep{zhang2022data, petravcek2021large} and a synthetic cave generator TT-Ribs~\citep{TT-ribs}, which has complex and irregular topography. From these two different categories of datasets, the agent is capable of learning to adapt to diverse environments, enhancing its ability to make accurate predictions in both regular and complex scenarios. This enables the agent to generalize its knowledge and improve its performance across a wide range of tasks.

\begin{figure}[!htbp]
    \centering
    \includegraphics[width=0.95\linewidth]{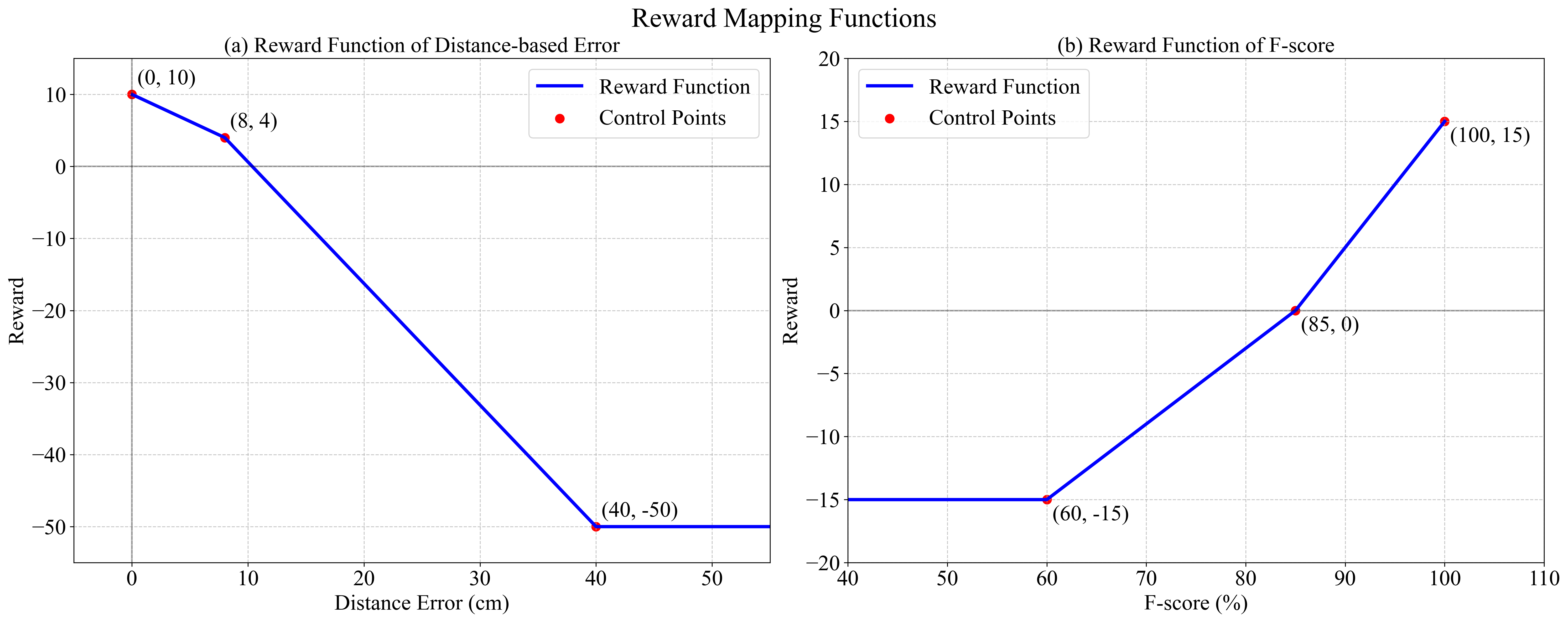}
    \caption{Piecewise linear reward mapping functions used in our learning and practice framework: (a) Conversion of distance-based errors (cm) to reward values, and (b) Conversion of F-score in percentages to reward values.}
    \label{fig:fenduanhanshu}
\end{figure}

For simulation settings, the UAV trajectories for simulation were pre-defined manually. Then the simulated laser scanning data will be processed into scanblocks. We utilize each scanblock as an interaction step with the environment, where the agent receives the current state (neural point cloud features), selects mapping parameters, and obtains rewards based on reconstruction quality metrics. For our experimental implementation, we designed piecewise linear reward conversion functions to transform the raw evaluation metrics into meaningful training signals, as illustrated in Fig.~\ref{fig:fenduanhanshu}. These functions provide appropriate gradient information throughout the performance range while emphasizing improvements in critical regions. For agent network training, we set the training hyperparameter as in Table~\ref{tab:training_params}.

The agent is trained using the Proximal Policy Optimization (PPO) algorithm implemented in Stable-Baselines3~\citep{raffin2021stable}. To stabilize training, we set a linear learning rate schedule that decays from \(3 \times 10^{-4}\) to \(3 \times 10^{-5}\) over the training period and employ gradient clipping with a threshold of $0.5$ and normalize both rewards and advantages. The agent's policy network was trained for approximately 1500 iterations until convergence, achieving stable and effective parameter selection performance across diverse unexposed environments. The higher weight assigned to F-score reward emphasizes the importance of balanced reconstruction quality.

\rmk{
Practice in simulation environments is a critical component of our reinforcement learning framework. By systematically exposing the agent to a wide variety of unexposed environments in simulation, it learns robust parameter adaptation strategies that generalize effectively to previously unseen real-world scenarios. This simulation-based approach provides several key advantages: it enables an immediate and accurate action-to-reward feedback loop that accelerates learning convergence; it allows comprehensive evaluation against ground truth data that is otherwise unobtainable in real-world unexposed settings; and it supports safe parameter exploration without operational risks, facilitating the testing of configurations that would be impractical or unsafe in field deployments.
}

\begin{table}[!htbp]
    \centering
    \caption{Training Parameters for PPO Algorithm}
    \label{tab:training_params}
    \begin{tabular}{lll} 
    \toprule
    \textbf{Parameter Category} & \textbf{Parameter Name} & \textbf{Value} \\
    \midrule
    \multirow{3}{*}{Training Setup} 
     & Number of Environments & 1 \\
     & Validation Interval & 10 \\
     & Batch Size & 20 \\
    \hline
    \multirow{5}{*}{PPO Parameters} 
     & Surrogate optimization epochs & 10 \\
     & Discount Factor ($\gamma$) & 0.6 \\
     & GAE Lambda ($\lambda$) & 0.95 \\
     & Entropy Coefficient & 0.0025 \\
     & Value Function Coefficient & 0.5 \\
    \hline
    \multirow{4}{*}{Reward Weights} 
     & $\lambda_1$ & 1 \\
     & $\lambda_2$ & 1 \\
     & $\lambda_3$ & 1 \\
     & $\lambda_4$ & 2 \\
    \bottomrule
    \end{tabular}
\end{table}

\subsection{Evaluation in Simulation Dataset}
To comprehensively evaluate the performance of our proposed method, we conducted experiments using synthetic data generated by MARSIM. For ground truth generation, we utilized TT-Ribs to create dense point clouds with a resolution of 0.1 meter. We set the height of the UAV at a constant height of 1.5 meters, and for a complete scanning of the scene, we set the scanning pattern as UA-MPC~\citep{li2025ua}. 

\begin{figure}[!htbp]
    \centering
    \includegraphics[width=\linewidth]{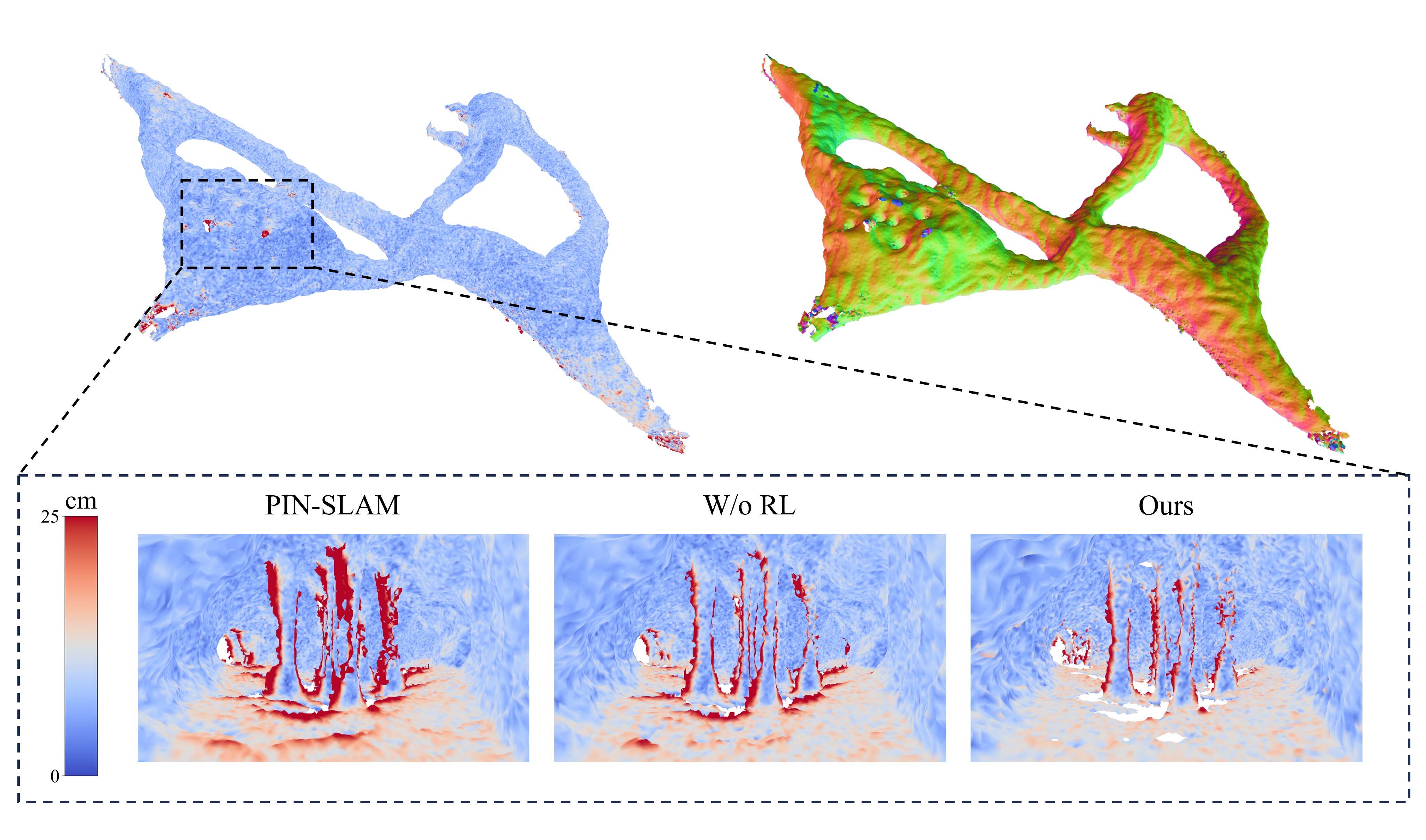}
    \caption{Visual comparison of different method in synthetic datasets.}
    \label{fig:synthetic}
\end{figure}

To systematically evaluate our approach, we conducted an ablation study across three synthetic unexposed environments generated in MARSIM. We used PIN-SLAM as the baseline and incrementally evaluated our contributions: first, implementing only the spatial-temporal geometry smoothing (Ours w/o RL), then adding the reinforcement learning-based meshing parameter optimization (Our complete system). Table~\ref{synthetic_quanti} presents the quantitative results using four metrics: accuracy, completeness, and Chamfer distance (all in centimeters, lower is better), and F-score (percentage, higher is better). To visually demonstrate the superiority of our proposed method, we present comparative visualizations between our reconstructed meshes (colored by surface normals) and the ground truth point cloud in Fig.~\ref{fig:synthetic}. The qualitative comparison clearly illustrates that our approach significantly reduces reconstruction artifacts and preserves geometric fidelity with higher accuracy. Specifically, our method demonstrates enhanced capability in recovering fine-grained structural details while maintaining global consistency, particularly in the karst formations and stalactite-stalagmite structures within the scene. These geometrically complex regions, where baseline methods typically struggle with noise and incomplete reconstruction, are precisely reconstructed by our approach with high fidelity to the original structure.

\begin{table*}[]
    \centering
    \caption{Quantitative evaluation on three synthetic unexposed scenes. We compare our method (both with and without RL component) against the PIN-SLAM. F-scores are calculated with a 15 cm threshold.}
    \label{synthetic_quanti}
    \begin{tabular*}{\textwidth}{@{\extracolsep{\fill}}llllll}
    \toprule
        Scene & Method & Accuracy(cm)$\downarrow$ & Completeness(cm)$\downarrow$ & Chamfer-L1(cm)$\downarrow$ & F-score(\%)$\uparrow$\\
    \midrule
        1 & PIN-SLAM        & 10.47 & 10.87 & 10.67 & 88.60 \\
          & Ours (w/o RL)   & 9.98 & 10.76 & 10.37 & 90.72 \\
          & Ours            & 9.55 & 10.35 & 9.95 & 92.53 \\
        \hline
        2 & PIN-SLAM        & 10.56 & 10.81 & 10.68 & 88.51 \\
          & Ours (w/o RL)   & 10.38 & 10.61 & 10.50 & 89.95 \\
          & Ours            & 9.74 & 10.20 & 9.97 & 92.36 \\
        \hline
        3 & PIN-SLAM        & 11.04 & 11.30 & 11.17 & 86.66 \\
          & Ours (w/o RL)   & 11.05 & 11.03 & 11.04 & 87.37 \\
          & Ours            & 9.95 & 10.43 & 10.19 & 91.08 \\
    \bottomrule
    \end{tabular*}
\end{table*}

\subsection{Evaluation in Open-Sourced Real-world Dataset}
\begin{figure*}[htbp]
    \centering
    \includegraphics[width=\linewidth]{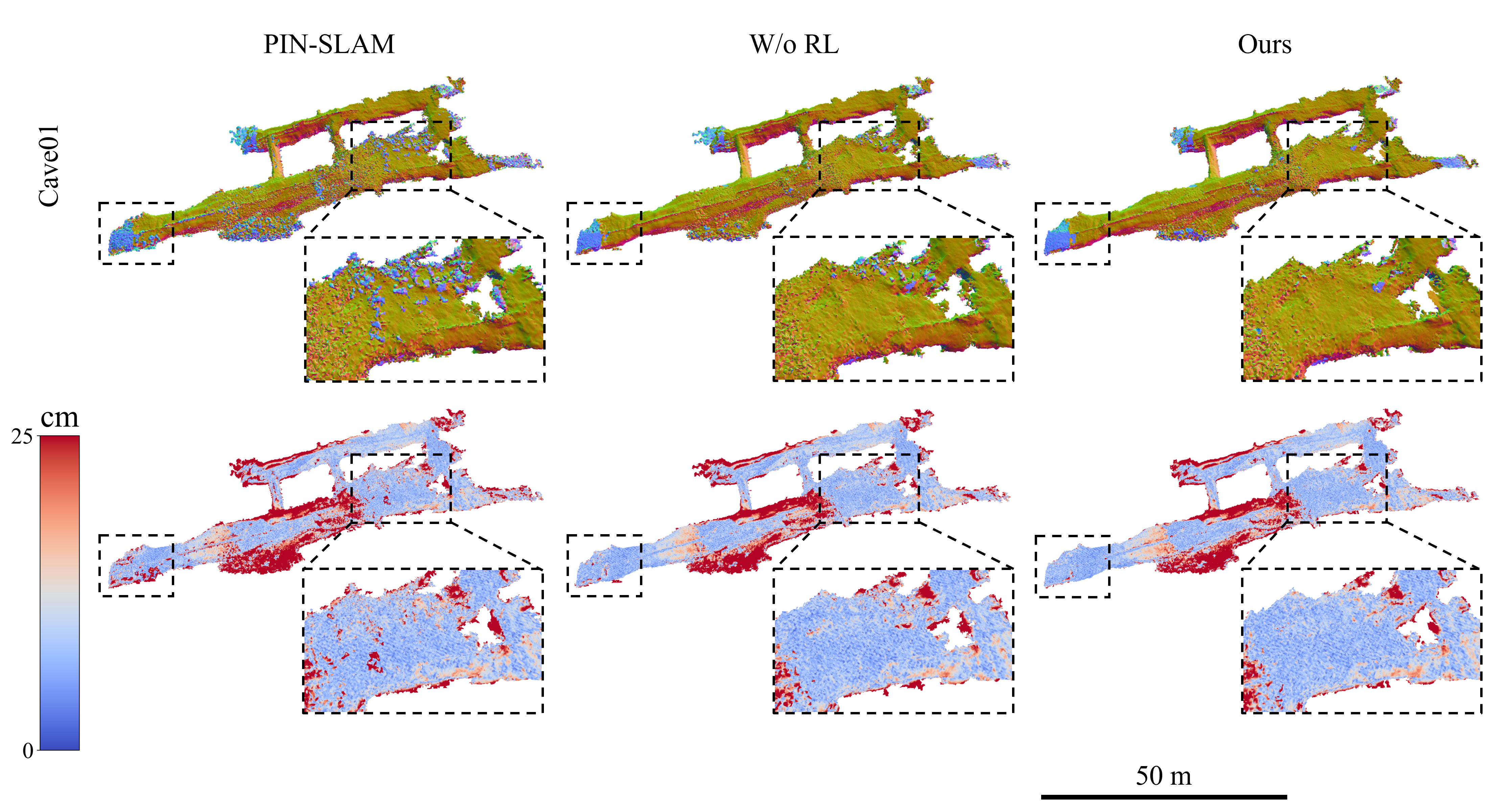}
    \caption{Comparative reconstruction results for the SuperLoc Cave01 dataset. The figure presents an incremental comparison among the baseline approach (PIN-SLAM), our method without reinforcement learning (w/o RL), and our complete method, with both the mesh reconstructions (colored by surface normals) and corresponding error maps visualized. Our complete method demonstrates superior geometric accuracy and detail preservation in complex cave environments.}
    \label{fig:cave01}
\end{figure*}

\begin{figure*}[htbp]
    \centering
    \includegraphics[width=\linewidth]{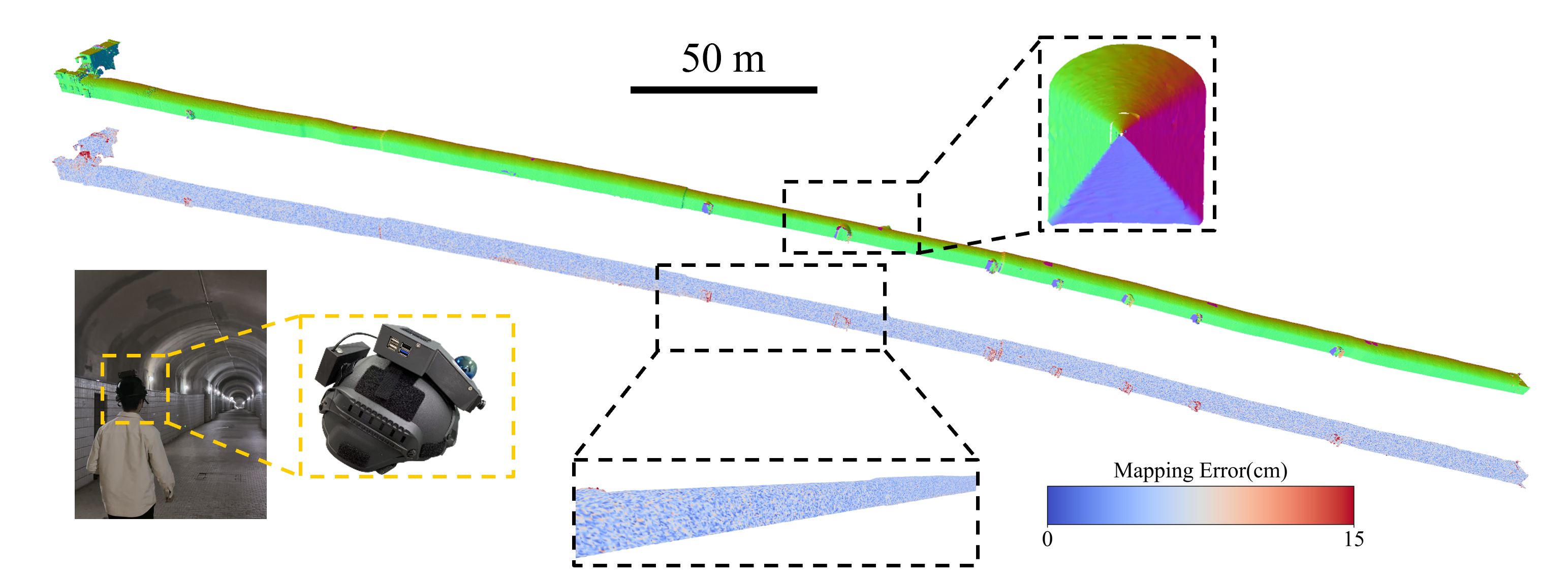}
    \caption{Qualitative reconstruction and error visualization results on the WHU-Helmet unexposed tunnel dataset. The visualization includes: (top) the reconstructed mesh generated by our method and colored according to surface normals; (middle) a comparison between our reconstruction and the ground truth point cloud with associated error metrics; and (bottom left) a schematic diagram illustrating the data acquisition process.}
    \label{fig:wudasuidao}
\end{figure*}

We evaluate our approach on three challenging real-world unexposed datasets: one tunnel environment from two cave environments from SuperLoc~\citep{zhao2024superloc} and WHU-Helmet~\citep{li2023whu}. The the SuperLoc datasets are acquired using a handheld device integrating a Velodyne VLP-16 LiDAR and an inertial measurement unit, while WHU-Helmet dataset is collected using a helmet-mounted mapping system equipped with a Livox Mid360 LiDAR scanner. Specifically, we select the the Cave01 and Cave02 sequences from SuperLoc and 3.1$\_$underground$\_$tunnel sequence from WHU-Helmet. For all three environments, terrestrial laser scanning (TLS) data is available as reference models for quantitative evaluation.
We conduct comprehensive comparisons against the baseline method and perform a detailed ablation study to validate each component's contribution. To ensure fair comparison, we maintain a voxel size of 15 cm for mesh generation. The comparative reconstruction quality in the SuperLoc Cave01 environment are presented in Fig.~\ref{fig:cave01}, while Fig.~\ref{fig:wudasuidao} illustrates the qualitative results for the WHU-Helmet tunnel environment. 
Visual inspection clearly demonstrates that our method produces significantly cleaner geometry with more accurate surface details compared to the baseline approaches, particularly in regions with complex structural features. This improvement can be attributed to our reinforcement learning network, which effectively mitigates potential artifacts cased by SDF label error, as highlighted in the regions marked by black boxes in Fig.~\ref{fig:cave01}. Furthermore, our reinforcement learning agent exhibits adaptive sampling behavior during different stages of reconstruction: initially increasing free-space sample density beyond conventional thresholds to establish mapping stability, then progressively reducing sampling rates below normal-guided sampling density to optimize computational efficiency while maintaining geometric accuracy. This dynamic parameter adaptation demonstrates the agent's learned policy for balancing reconstruction quality and processing speed.

For the quantitative evaluation, we use the 3D reconstruction metrics we have mentioned before, namely accuracy, completeness, Chamfer-L1 distance, and F-score, which could describe the reconstruction quality in a comprehensive way. We calculate the evaluation metrics as in Table~\ref{tab:superloc} and Table~\ref{tab:whu-helmet}.

\begin{table*}[htbp]
\centering
\caption{Quantitative results of SuperLoc dataset. We show the F-score in \% with a 30 cm error threshold.}
\label{tab:superloc}
\begin{tabular*}{\textwidth}{@{\extracolsep{\fill}}lcccccccc}
\toprule
& \multicolumn{4}{c}{cave01} & \multicolumn{4}{c}{cave04} \\
\cmidrule(lr){2-5} \cmidrule(lr){6-9}
{Method} & {Acc.(cm)$\downarrow$} & {Comp.(cm)$\downarrow$} & {C-L1(cm)$\downarrow$} & {F-score(\%)$\uparrow$}
                & {Acc.(cm)$\downarrow$} & {Comp.(cm)$\downarrow$} & {C-L1(cm)$\downarrow$} & {F-score(\%)$\uparrow$} \\
\midrule
PIN-SLAM      & 15.43 & 14.48 & 14.95  & 89.94 & 19.50 & 9.53 & 14.51  & 87.71 \\
Ours (w/o RL) & 14.51 & 12.16 & 13.33  & 91.20 & 19.17 & 9.58 & 14.37  & 88.30  \\
Ours          & 12.80 & 12.44 & 12.62  & 93.50 & 17.98 & 9.66 & 13.82  & 90.36 \\
\bottomrule
\end{tabular*}
\end{table*}

\begin{table}[htbp]
    \centering
    \caption{Quantitative results of WHU-Helmet dataset. We show the Accuracy(Acc.), Completeness(Comp.) and C-L1 in centimeters, while F-score in \% with a 10 cm error threshold.}
    \label{tab:whu-helmet}
    \begin{tabular}{lllll}
    \toprule
        Method & Acc.$\downarrow$ & Comp.$\downarrow$ & C-L1$\downarrow$ & F-score$\uparrow$\\
    \midrule
        PIN-SLAM       & 7.55 & 5.28 & 6.42 & 86.06 \\
        Ours (w/o RL)  & 7.34 & 5.21 & 6.27 & 87.46 \\
        Ours           & 6.28 & 4.89 & 5.59 & 91.40 \\
    \bottomrule
    \end{tabular}
\end{table}

\subsection{Evaluation on In-house Helmet-based System}

\begin{figure}[htbp]
    \centering
    \includegraphics[width=0.95\linewidth]{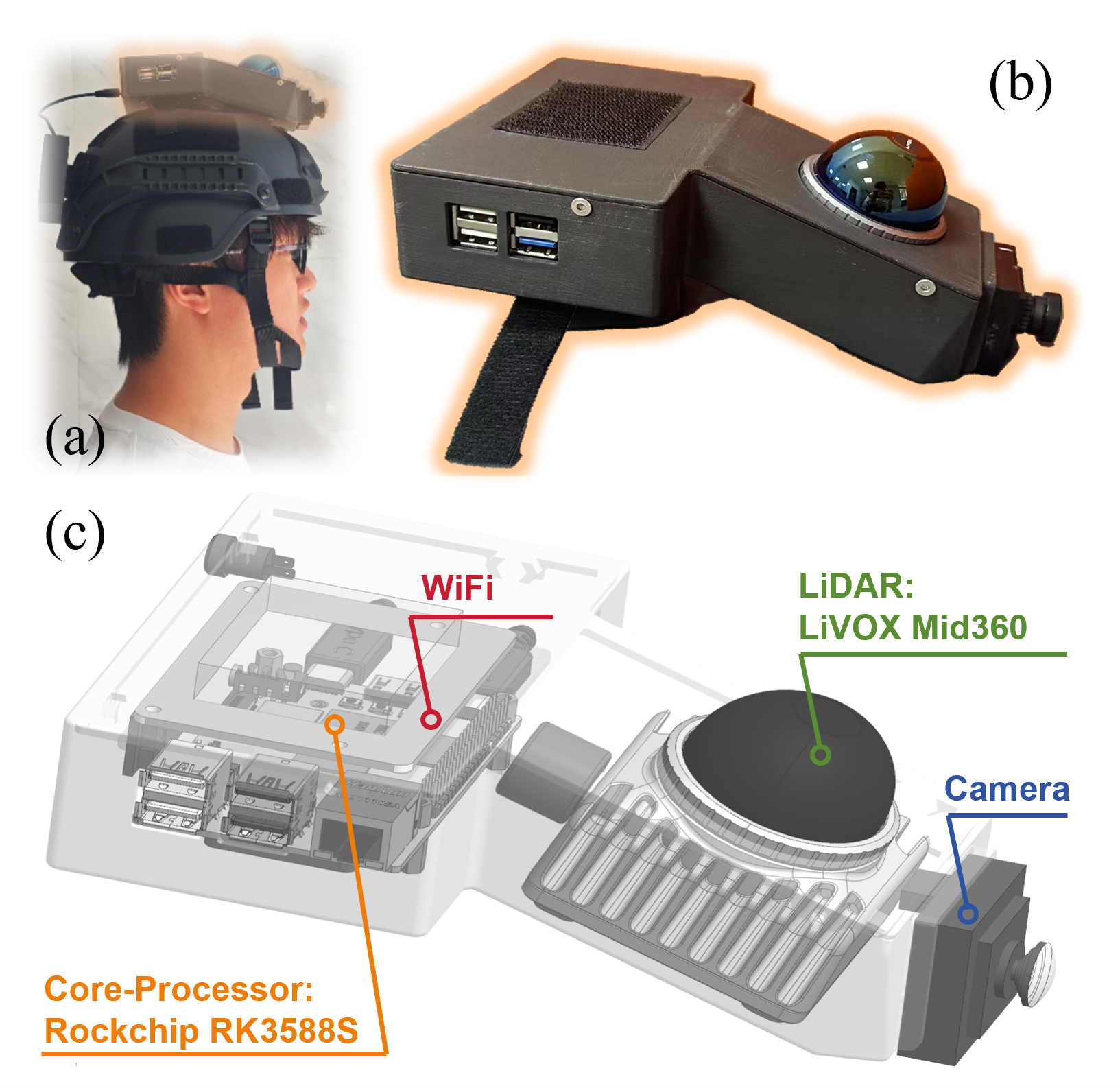}
    \caption{An in-house helmet-based mapping system. (a) Deployment illustration showing operator wearing the integrated helmet system during experiment; (b) Close-up view of the core mapping equipment; (c)Hardware architecture diagram highlighting key components including Livox Mid360 LiDAR, IMU, camera, WiFi and RK3588S processor.}
    \label{fig:inhousehardware}
\end{figure}

Inspired by the WHU-Helmet design, we developed a lightweight, helmet-integrated meshing system for the unexposed environment. The system architecture, illustrated in Fig.~\ref{fig:inhousehardware}, consists of a Livox Mid360 LiDAR for primary sensing, a high-precision IMU for motion tracking, and a supplementary camera for visual data acquisition. We use a modern CPU RK3588S for some lightweight task processing. A critical feature of our implementation is the precise temporal synchronization between all sensors, ensuring accurate spatial registration of acquired measurements. The system is in small size and with a low weight, facilitates single-person data collection.

\begin{figure}[htbp]
    \centering
    \includegraphics[width=0.95\linewidth]{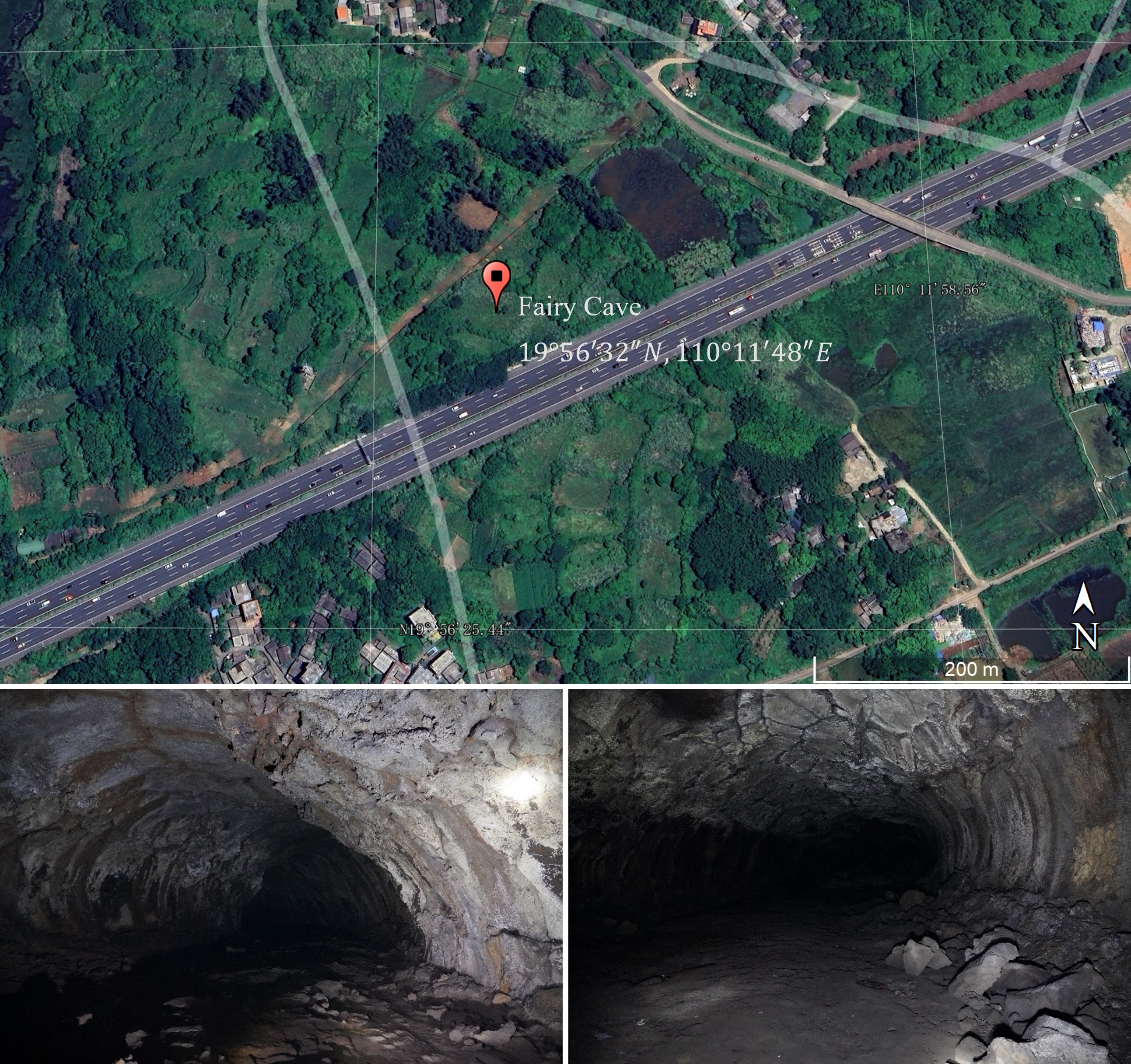}
    \caption{Xianren Lava Tube field test site: (top) geographical location in satellite imagery, and (bottom) challenging interior environment characterized by complex geological formations, including irregular surfaces, rockfall accumulations, and natural collapse features that present significant challenges for 3D reconstruction.}
    \label{fig:xianrendong_pic}
\end{figure}

\begin{figure*}[htbp]
    \centering
    \includegraphics[width=0.95\linewidth]{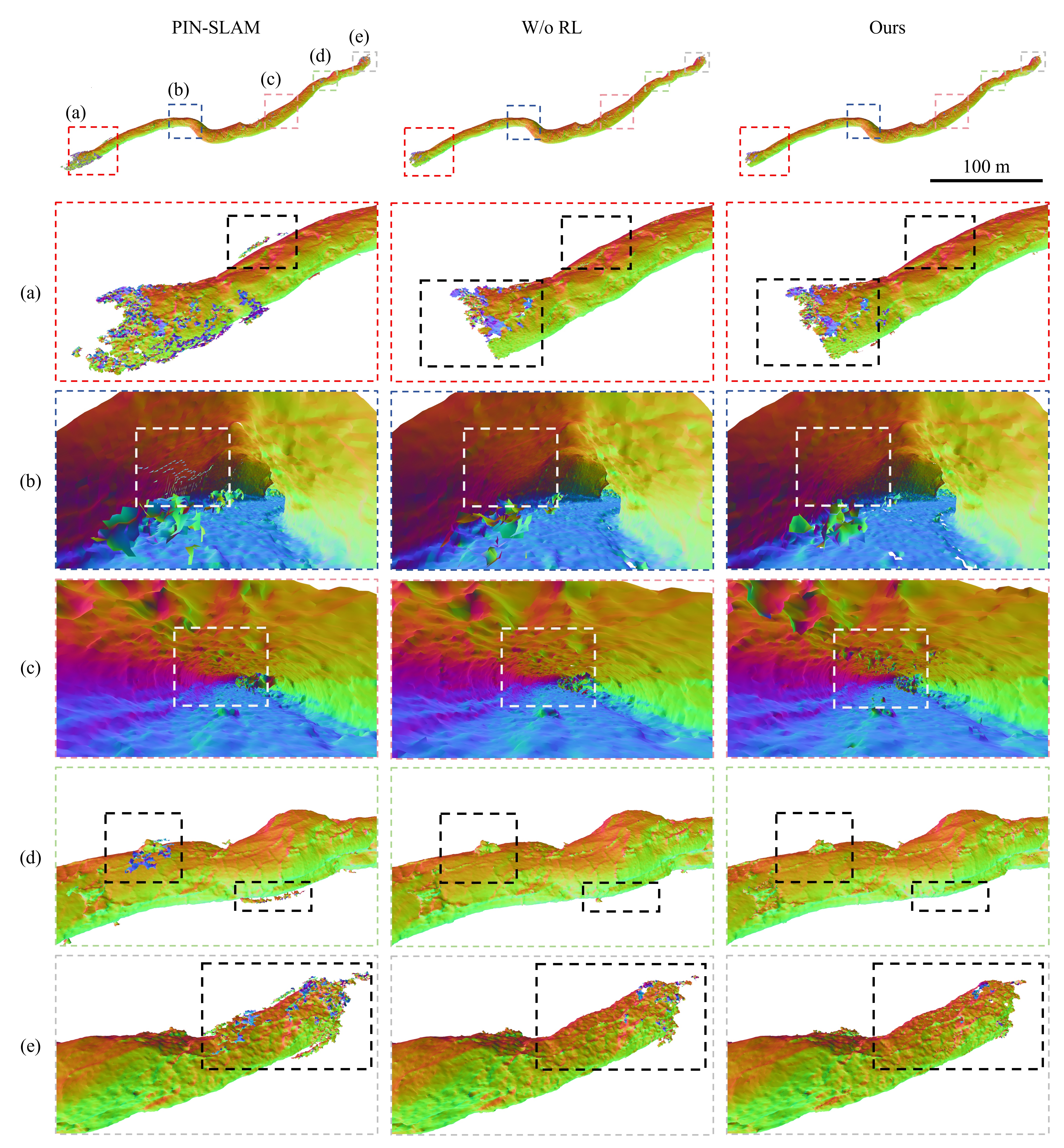}
    \caption{Qualitative comparison of reconstruction results for the Xianren Lava Tube. Significant differences between our method and the PIN-SLAM baseline are highlighted by bounding boxes. Our approach produces cleaner geometry with better preservation of structural details, effectively reducing common reconstruction artifacts and avoiding over-smoothing that typically occurs in complex unexposed environments with irregular surface features.}
    \label{fig:xianrendong}
\end{figure*}

\begin{figure}[htbp]
    \centering
    \includegraphics[width=0.95\linewidth]{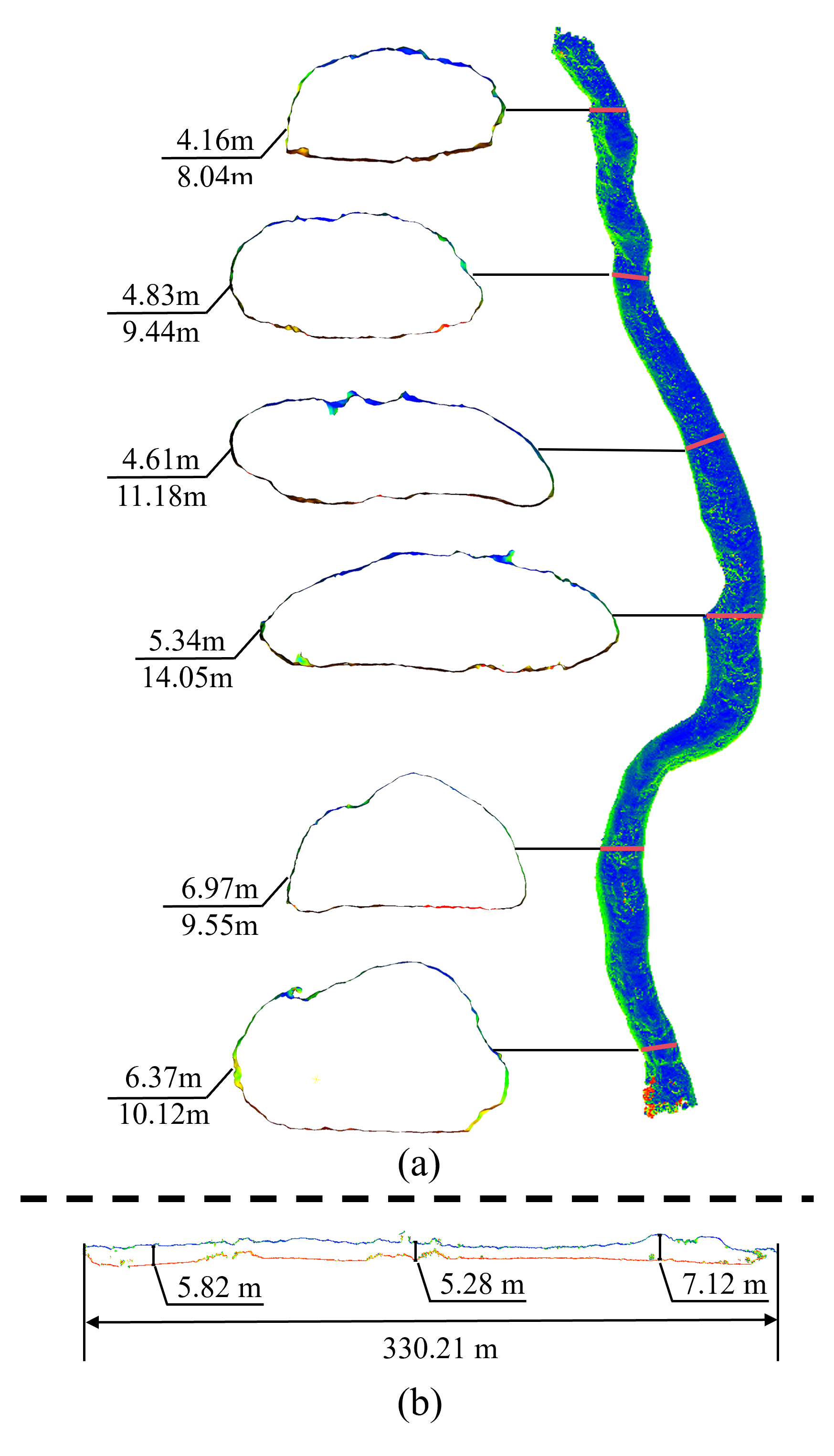}
    \caption{A cross-sectional analysis of Xianren Lava Tube: (a) Reconstructed mesh with horizontal cross-sections, color-coded by elevation gradient. The upper and lower numbers associated with each horizontal line represent the width and height of lava tube, respectively; (b) Vertical section along the 330.21m lava tube, where each marker indicates the ceiling height at the corresponding position.}
    \label{fig:jiemian}
\end{figure}

To evaluate the system's performance in challenging real-world conditions, we conducted an extensive field test in Xianren Lava Tube, a complex volcanic cave system located in Hainan Province, China. This environment presents numerous challenges for 3D meshing, including a continuously curved tunnel structure, substantial rockfall accumulations, and multiple small-scale collapses creating complex surface irregularities, as illustrated in Fig.~\ref{fig:xianrendong_pic}. Data acquisition was performed through a complete 300-meter round-trip trajectory, with a total scanning time of approximately 15 minutes. Since terrestrial laser scanning reference data is not available for this site, we present a qualitative comparative analysis in Fig.~\ref{fig:xianrendong}. Through incremental comparison between our method and the baseline PIN-SLAM reconstruction, the results demonstrate that our approach produces significantly cleaner geometry with more accurate representation of complex surface features, particularly evident in regions with challenging geological formations. 
Furthermore, we conducted a rigorous mesh-to-mesh distance analysis comparing our reconstruction against traditional methods using professional software Leica Cyclone 3DR. The quantitative evaluation demonstrates that 94.8\% of the points in our reconstructed mesh fall within the error threshold of $\pm5$cm, indicating superior geometric accuracy and consistency throughout the complex structure.
To provide a more comprehensive visualization and structural analysis of the reconstructed cave system, we present detailed cross-sectional views in Fig.~\ref{fig:jiemian}, including both longitudinal (vertical) and transverse (horizontal) sections through the Xianren Lava Tube. These sectional analyses are particularly valuable for geological studies and structural assessment, as they reveal internal morphological features such as ceiling height variations, passage width distributions, and volumetric characteristics that would otherwise be difficult to quantify from external 3D views alone. The cross-sectional analysis demonstrates our method's ability to maintain geometric accuracy and continuity throughout the complex tunnel structure, even in regions with significant morphological changes or geological irregularities. Such sectional visualizations are essential tools for speleologists, geologists, and civil engineers requiring precise dimensional data for stability assessment, preservation planning, or comparative morphological studies.

\subsection{Application in Tunnel Construction}

To validate the practical application of our method for the unexposed environment, especially in tunnel over-excavation and under-excavation analysis, we compare the volumetric calculations between our real-time mesh reconstruction and ground truth point cloud. We utilize professional software Leica Cyclone 3DR for volumetric analysis, comparing the excavation volumes derived from our method against the ground truth results. We also include volumetric measurements from the baseline method to further demonstrate our approach's effectiveness.

\begin{figure}[htbp]
    \centering
    \includegraphics[width=0.95\linewidth]{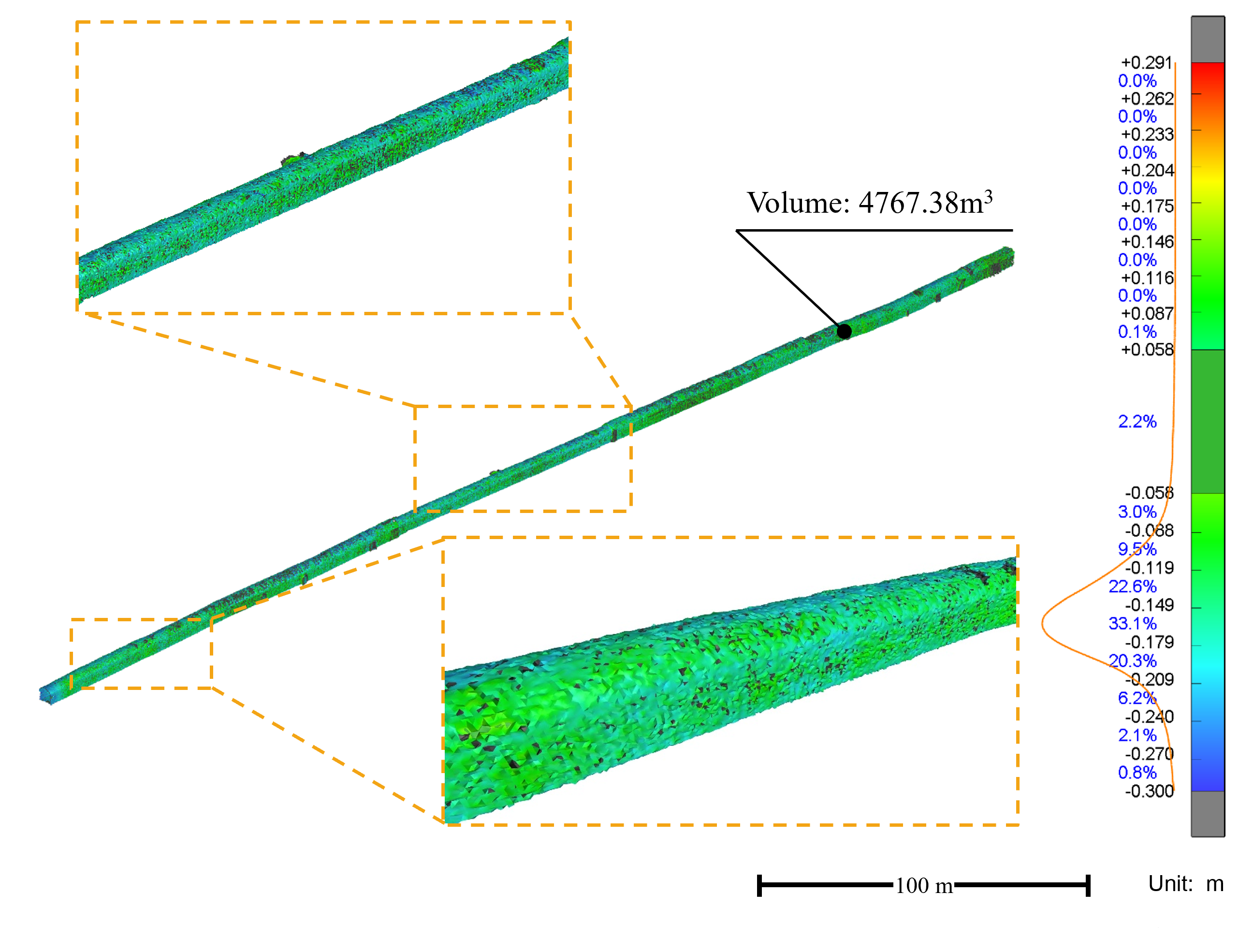}
    \caption{Visualization of over-excavation and under-excavation analysis in tunnel construction, showing spatial error distribution between reconstructed models and ground truth.}
    \label{fig:excavation}
\end{figure}

The meshing results and the corresponding error map are illustrated in Fig.~\ref{fig:excavation}. Our volumetric analysis reveals significant differences in accuracy: the ground truth point cloud yields an excavation volume of \(4714.63\,m^3\), our method produces \(4767.38\,m^3\), while the baseline results in \(4825.22\,m^3\). The baseline consistently overestimates the volume compared to both our method and ground truth, which can be attributed to its use of training labels with projection errors during SDF field computation, leading to volume inflation. In contrast, our approach achieves more accurate volume estimation with a relative error of only \(1.11\%\). This significant improvement is attributed to two key innovations in our approach: (1) dynamic adjustment of meshing parameters through reinforcement learning, which allows the system to adapt reconstruction settings based on local geometry complexity, and (2) more accurate geometry sampling based on our spatially smoothed point cloud, which reduces noise while preserving structural details. Unlike the baseline method that employs fixed parameters and relies on labels with inherent projection errors, our system adaptively optimizes the reconstruction process according to specific scene characteristics and utilizes non-projected geometry for sampling. This adaptive strategy substantially reduces volumetric estimation errors in tunnel excavation analysis, making our approach particularly valuable for civil engineering applications requiring high measurement precision.

\section{Conclusion}
\label{Conslusion}
This paper proposed ARMOR, an adaptive meshing system with reinforcement learning for real-time 3D monitoring in unexposed scenes. Specifically, we first implemented a spatial-temporal normal smoothing mechanism that leverages multi-view consistency to enhance geometry coherence and eliminate projection-based SDF label errors. Then, a reinforcement learning framework was developed to adaptively optimize meshing parameters based on local scene characteristics, eliminating the need for manual tuning while ensuring consistent reconstruction quality across diverse environments. Comprehensive experiments conducted across more than 3000 meters of underground environments, including tunnels, caves, and lava tubes, demonstrated that ARMOR significantly improves reconstruction accuracy by 3.96\% while maintaining real-time performance. In future work, we plan to explore the integration of our system with autonomous robotic platforms for inspection tasks, and extend our approach to handle dynamic elements in underground construction sites, such as moving equipment and temporary structures.


\printcredits

\section{Declaration of interests}
The authors declare that they have no known competing financial interests or personal relationships that could have appeared to influence the work reported in this paper.

\section{Acknowledgment}
This study was jointly supported by the National Natural Science Foundation Project(No.42201477,No.42130105).

\appendix
\section{Auxiliary Variable Optimization for $L_0$ Minimization}

To address the \(L_0\) minimization problem in Eq. (\ref{min_l0norm}), we introduce an auxiliary variable \(\boldsymbol{\zeta}\) to facilitate the solution:
\begin{equation}
    \label{min_l0_aux}
    F = \min _{\mathbf{n},\left|\mathbf{n}\right|=1,\boldsymbol{\zeta}}1-\mathbf{n}^{\top}\mathbf{\hat{n}}+\beta\lvert \mathbf{D}(\mathbf{n})-\boldsymbol{\zeta}\rvert^2+\eta\lvert\boldsymbol{\zeta}\rvert_0,
\end{equation}
where \(\beta\) controls the convergence speed. Next, we initialize \(\mathbf{n}\) with an initial value \(\mathbf{\hat{n}}\) and fix it as a constant, then solve for \(\boldsymbol{\zeta}\) as shown below:
\begin{equation}
    \label{iter_1}
    F = \min _{\boldsymbol{\zeta}} \beta\lvert \mathbf{D}(\mathbf{n})-\boldsymbol{\zeta}\rvert^{2}+\eta\lvert\boldsymbol{\zeta}\rvert_{0}.
\end{equation}
The solution to this problem is given as follows:
\begin{equation}
    \label{given}
    \boldsymbol{\zeta}_i =
        \begin{cases}
            0 & \text{if } \dfrac{\eta}{\beta} > \lvert \mathbf{D}(\mathbf{n}_i) \rvert^2, \\
            \mathbf{D}(\mathbf{n}_i) & \text{otherwise}.
        \end{cases}
\end{equation}
Subsequently, fixing \(\boldsymbol{\zeta}\) as a constant, solving Eq. (\ref{min_l0_aux}):
\begin{equation}
    \label{iter_2}
    F = \min _{\mathbf{n},\left|\mathbf{n}\right|=1}1-\mathbf{n}^{\top}\mathbf{\hat{n}}+\beta|\mathbf{D}(\mathbf{n})-\boldsymbol{\zeta}|^2.
\end{equation}
This equation is quadratic with a global minimum, and to speed up solving for the smoothed normal vector, we give the explicit solution as:
\begin{equation}
    \mathbf{n}_i=\frac{\hat{\mathbf{n}}_i+\beta \cdot \sum_{j}^{k}(\mathbf{n}_{\mathbf{N}(i,j)}+\boldsymbol{\zeta}_{ik+j})}{\beta \cdot k +1}.
\end{equation}
During the optimization process, Eq. (\ref{iter_1}) and Eq. (\ref{iter_2}) are alternately updated in each iteration, with \(\beta \leftarrow 2\beta\) after each loop. This iterative approach ultimately drives \(\mathbf{D}(\mathbf{n})\) to converge towards \(\boldsymbol{\zeta}\).

\bibliographystyle{cas-model2-names}

\bibliography{reference}

\end{sloppypar}
\end{document}